\documentclass[sigconf]{acmart}
\AtBeginDocument{%
  }

\copyrightyear{2026}
\acmYear{2026}

\setcopyright{cc}
\setcctype{by}

\acmConference[MM '26]
  {Proceedings of the 34th ACM International Conference on Multimedia}
  {November 10--14, 2026}
  {Rio de Janeiro, Brazil}

\acmBooktitle{Proceedings of the 34th ACM International Conference on
Multimedia (MM '26), November 10--14, 2026, Rio de Janeiro, Brazil}

\acmISBN{979-8-4007-2213-4/2026/11}

\acmDOI{10.1145/3767308.3835527}

\usepackage{multirow}
\usepackage{colortbl}
\usepackage{cuted}

\begin{document}
\title[TR-GS for Sparse-View CT]
{TR-GS: High-Fidelity Sparse-View CT Volumetric Rendering via
t-Distribution Gaussian Splatting and Ray-Confidence Modeling}

\author{Zedong Xiao}
\affiliation{%
  \institution{Shenzhen University}
  \city{Shenzhen}
  \country{China}}
\email{2023111035@email.szu.edu.cn}

\author{Yiren Wang}
\affiliation{%
  \institution{Shenzhen University}
  \city{Shenzhen}
  \country{China}}
\email{2510232069@mails.szu.edu.cn}

\author{Zhou Liu}
\affiliation{%
  \institution{Guangdong Laboratory of Artificial Intelligence and Digital Economy (SZ)}
  \city{Shenzhen}
  \country{China}}
\email{liuzhou@gml.ac.cn}

\author{Xiaolin Liu}
\affiliation{%
  \institution{Shenzhen University}
  \city{Shenzhen}
  \country{China}}
\email{liuxiaolinszu@163.com}

\author{Zhangji Lu}
\affiliation{%
  \institution{Shenzhen University}
  \city{Shenzhen}
  \country{China}}
\email{zjlu@szu.edu.cn}

\renewcommand{\shortauthors}{Xiao et al.}

\begin{abstract}
High-fidelity 3D medical visualization supports applications such as
clinical assessment and surgical planning. Sparse-view computed
tomography (CT) can reduce projection requirements and associated
radiation exposure, but limited observations may introduce structural
artifacts and reconstruction uncertainty. Although 3D Gaussian
Splatting (3DGS) provides an efficient explicit representation for
volumetric rendering, existing CT methods based on standard Gaussian
primitives may be sensitive to unreliable observations under
sparse-view acquisition.

We present TR-GS, a Gaussian-splatting framework for sparse-view CT
volumetric rendering. TR-GS replaces standard Gaussian primitives with
projectable Student's \(t\)-distribution primitives and introduces a
ray-confidence model that regulates their degrees of freedom according
to local ray observability. Confidence-guided 3D wavelet regularization
is further used to balance high-frequency detail preservation and noise
suppression.

Experiments on synthetic and real-world datasets show that TR-GS
improves over representative baselines in most evaluated settings and
remains competitive in the remaining cases. The resulting volumetric
representations may support downstream medical multimedia applications,
including XR-based visualization and interactive clinical rendering.
\end{abstract}


\begin{teaserfigure}
  \includegraphics[width=\textwidth]{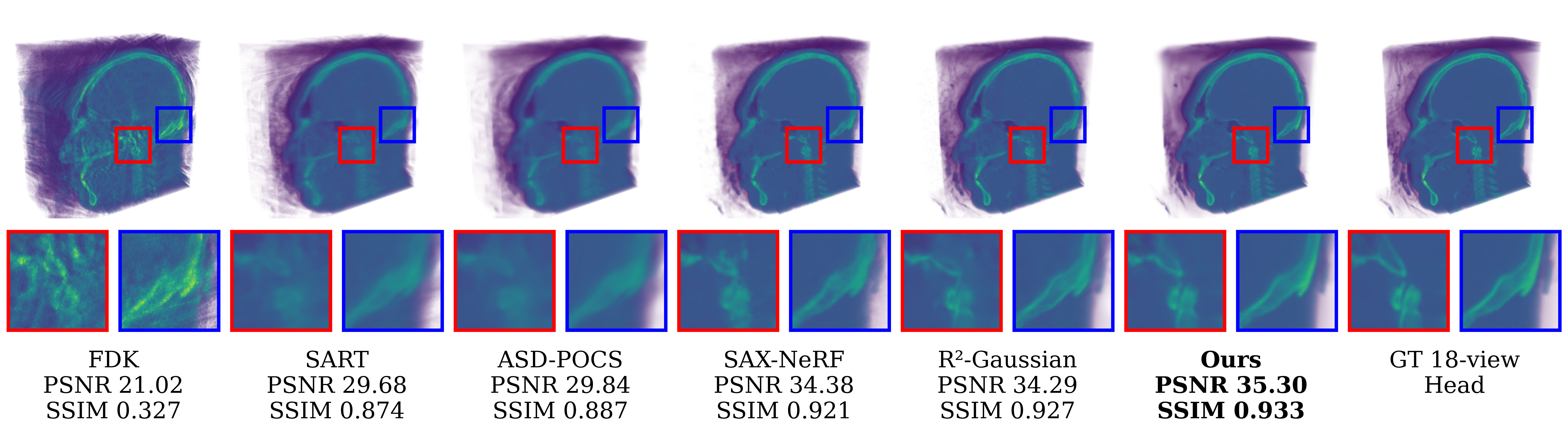}
  \caption{Sparse-view CT reconstruction comparison at 18 views. In this example, TR-GS reconstructs clear anatomical structures and obtains a PSNR of 35.30~dB and an SSIM of 0.933.}
  \Description{Comparison of sparse-view CT reconstruction results from different methods at 18 views. TR-GS produces clearly visible anatomical structures with reduced artifacts.}
  \label{fig:teaser}
\end{teaserfigure}

\maketitle
\section{Introduction}
Computed tomography (CT) is a fundamental imaging technique for non-invasive visualization of internal structures and has been widely used in medical diagnosis, biological imaging, and industrial inspection~\cite{Hounsfield1980,Cormack1963,KakSlaney2001,DeChiffre2014}. In clinical applications, CT-based volumetric reconstruction is closely related to medical visualization, surgical planning, and image-guided intervention. However, high-quality CT imaging usually requires a large number of X-ray projections, leading to increased radiation exposure and potential long-term health risks~\cite{Mostafapour2024,Koch2024,brenner2007computed}. Sparse-view CT aims to alleviate this issue by reconstructing volumetric images from a limited number of projections, thereby reducing radiation dose while preserving diagnostically useful anatomical information~\cite{DavidOlawade2025,Shieh2019,Keall2004,mccollough2009strategies}.

Despite its practical value, sparse-view CT reconstruction remains a highly ill-posed inverse problem. With insufficient projection views, conventional analytical reconstruction methods such as FBP~\cite{RamachandranLakshminarayanan1971} and FDK~\cite{FeldkampDavisKress1984} often suffer from severe streak artifacts and structural distortion. Iterative approaches, including SART~\cite{AndersenKak1984} and ASD-POCS~\cite{SidkyPan2008}, can partially improve reconstruction quality by incorporating handcrafted priors, but they typically require expensive optimization and may over-smooth fine details under aggressive sparsity. To overcome these limitations, a large body of recent work has explored learning-based sparse-view CT reconstruction, including adversarial learning~\cite{wolterink2017generative}, implicit neural representations, and diffusion-based priors~\cite{Ying2019,Lantz2024,Chung2023,Liu2020,Liu2023,Anirudh2018}. However, many supervised methods depend on large paired datasets and may generalize poorly to unseen anatomies~\cite{Lantz2024,Chung2023}.

To reduce the reliance on external supervision, self-supervised neural rendering methods have recently been introduced into CT reconstruction. Inspired by NeRF~\cite{Mildenhall2020}, methods such as IntraTomo~\cite{Zang2021}, NAF~\cite{Zha2022}, and SAX-NeRF~\cite{Cai2024} reconstruct volumetric attenuation fields directly from sparse projections of a single object or patient. These approaches have demonstrated encouraging performance by optimizing continuous scene representations with differentiable volume rendering. Nevertheless, NeRF-based methods require repeated neural network evaluation along sampled rays, resulting in high computational cost and slow convergence, especially for high-resolution CT volumes~\cite{Zha2022,Cai2024}.

Recently, 3D Gaussian Splatting (3DGS) has emerged as an efficient explicit scene representation for differentiable rendering~\cite{Kerbl2023}. Owing to its explicit primitive-based formulation and fast rasterization, Gaussian splatting has been extended from natural scene rendering to tomographic imaging and X-ray reconstruction~\cite{cai2024radiative,Nikolakakis2024,Lin2023SparseView,Zha2024R2Gaussian}. In particular, R\(^2\)-Gaussian~\cite{Zha2024R2Gaussian} shows that explicit Gaussian-based representations can significantly improve the efficiency of sparse-view tomographic reconstruction while maintaining competitive reconstruction quality. However, existing Gaussian-splatting-based CT methods still rely on standard Gaussian primitives. Under sparse-view acquisition, the reconstruction process is often affected by outliers, missing constraints, and local uncertainty caused by incomplete angular coverage. Since Gaussian distributions are thin-tailed, they are less robust to these unreliable observations, which can lead to unstable primitive optimization, residual artifacts, and loss of structural detail in the reconstructed volume.

To address this limitation, we propose TR-GS, a sparse-view CT reconstruction framework based on t-distribution radiative splatting. Instead of modeling each primitive with a Gaussian kernel, TR-GS adopts a heavy-tailed Student's t-distribution to improve robustness against sparse-view ambiguity and outlier contamination. Furthermore, we introduce a ray-confidence modeling strategy to estimate the reliability of each primitive according to the geometric support provided by the observed X-ray rays. This confidence is used to adaptively regulate the primitive distribution during optimization, allowing well-constrained regions to preserve precision while encouraging under-constrained regions to remain robust. We further incorporate a confidence-guided 3D wavelet regularization to suppress sparse-view artifacts and noise while preserving high-frequency anatomical structures that are critical for CT visualization. The main contributions of this work are summarized as follows:
\begin{itemize}
    \item We formulate Student's \(t\)-distributions as explicit, projectable radiative primitives for sparse-view CT, and couple their degrees of freedom with local ray observability through a ray-confidence model.
    \item We develop a ray-confidence model that adaptively regulates the degrees of freedom for each rendering primitive, and introduce a 3D wavelet regularization guided by the ray-confidence model to preserve high-frequency clinical details.
    \item Extensive experiments on synthetic and real-world datasets show that TR-GS improves over representative baselines in most evaluated settings and remains competitive in the remaining cases, supporting its value for high-fidelity medical visualization and sparse-view CT rendering.
    \item The resulting artifact-reduced volumetric representations may support medical multimedia applications such as XR-based visualization and interactive clinical rendering~\cite{barsom2016systematic}.
\end{itemize}

Our implementation is publicly available at \url{https://github.com/zd-X/TR-GS}.

\section{Related Work}

In medical imaging literature, the process of inferring 3D volumetric data from 2D projections is commonly referred to as CT reconstruction. In computer graphics and multimedia domains, the same process of synthesizing novel views from volumetric representations is termed volumetric rendering. Throughout this paper, we use these terms interchangeably to bridge the medical imaging and multimedia communities.

\subsection{Sparse-View CT Reconstruction}
CT reconstruction has been extensively studied to reduce radiation dose while maintaining acceptable image quality~\cite{Mostafapour2024,Koch2024,DavidOlawade2025,Shieh2019,Keall2004}. Classical methods mainly include analytical and iterative reconstruction algorithms. Analytical methods such as FBP~\cite{RamachandranLakshminarayanan1971} and FDK~\cite{FeldkampDavisKress1984}, derived from the Radon transform~\cite{radon1986determination}, are computationally efficient but are highly sensitive to view sparsity and usually produce strong streak artifacts when projections are insufficient~\cite{KakSlaney2001,SidkyPan2008}. They are derived from the Radon transform theory~\cite{radon1986determination}. Iterative reconstruction methods, including SART~\cite{AndersenKak1984} and ASD-POCS~\cite{SidkyPan2008,sidky2006accurate}, improve reconstruction quality by incorporating prior constraints into the inverse problem. Nevertheless, these methods often rely on carefully designed regularization terms and may lose fine structural details in highly sparse settings.

With the development of deep learning, data-driven reconstruction methods have achieved significant progress in sparse-view CT. Earlier studies used adversarial learning and convolutional networks to recover CT volumes from limited projections or degraded reconstructions~\cite{Ying2019,Liu2020}. More recent methods have explored diffusion priors and model-based generative frameworks to further improve reconstruction quality in sparse or limited-angle scenarios~\cite{Chung2023,Liu2023,Anirudh2018}. Although these methods can produce strong quantitative performance, many of them require large-scale paired training data and may suffer from limited generalization to unseen domains~\cite{Lantz2024,Chung2023}.

\subsection{Neural Fields for CT Reconstruction}
Neural field methods have recently provided a self-supervised alternative for tomographic reconstruction by directly optimizing a continuous volumetric representation from acquired projections. NeRF~\cite{Mildenhall2020} first demonstrated the effectiveness of neural scene representations for view synthesis, and subsequent work extended this idea to tomography and X-ray imaging~\cite{Zang2021,Zha2022,Cai2024,Lin2023,Ruckert2022}. IntraTomo~\cite{Zang2021} formulates tomographic reconstruction as self-supervised learning via sinogram synthesis and prediction. NAF~\cite{Zha2022} introduces neural attenuation fields for sparse-view cone-beam CT reconstruction and improves efficiency with hash-based encodings~\cite{Muller2022}. SAX-NeRF~\cite{Cai2024} exploits X-ray structure information through line-segment-based transformers and structure-aware ray sampling. NeAT~\cite{Ruckert2022} proposes adaptive neural representations for tomography.

These methods reduce the need for external annotations and can flexibly adapt to instance-specific reconstruction. However, because they repeatedly evaluate neural networks along many sampled points on each ray, neural-field-based methods~\cite{Zha2022,Cai2024,Mildenhall2020} are usually computationally expensive and slow to optimize, especially for large-scale 3D medical volumes.

\subsection{Gaussian Splatting for CT Reconstruction}
3D Gaussian Splatting (3DGS)~\cite{Kerbl2023} has recently become an efficient alternative to implicit neural rendering due to its explicit primitive representation and real-time differentiable rasterization. Beyond natural scene rendering, Gaussian splatting has been rapidly extended to 3D reconstruction, surface representation, generation, and SLAM~\cite{GuedonLepetit2024,Lu2024,Tang2024,Matsuki2024}. This progress has motivated its application to radiographic imaging and CT reconstruction~\cite{cai2024radiative,Nikolakakis2024,Lin2023SparseView,Zha2024R2Gaussian}.

X-Gaussian~\cite{cai2024radiative} introduces Gaussian-based radiative primitives for X-ray rendering and redesigns the point cloud model to exclude view-direction influence, inspired by the isotropic nature of X-ray imaging. GASPCT~\cite{Nikolakakis2024} uses Gaussian splatting for novel CT projection view synthesis. Sparse-view CT reconstruction with 3D Gaussian volumetric representation further explores explicit Gaussian representations for density estimation~\cite{Lin2023SparseView}. R\(^2\)-Gaussian~\cite{Zha2024R2Gaussian} identifies an integration bias in conventional 3DGS when applied to tomography and proposes a rectified radiative Gaussian splatting framework for sparse-view reconstruction. These works show that Gaussian-splatting-based reconstruction is much more efficient than NeRF-style volumetric rendering while maintaining strong performance.

\begin{figure*}[t]
    \centering
    \includegraphics[width=\textwidth]{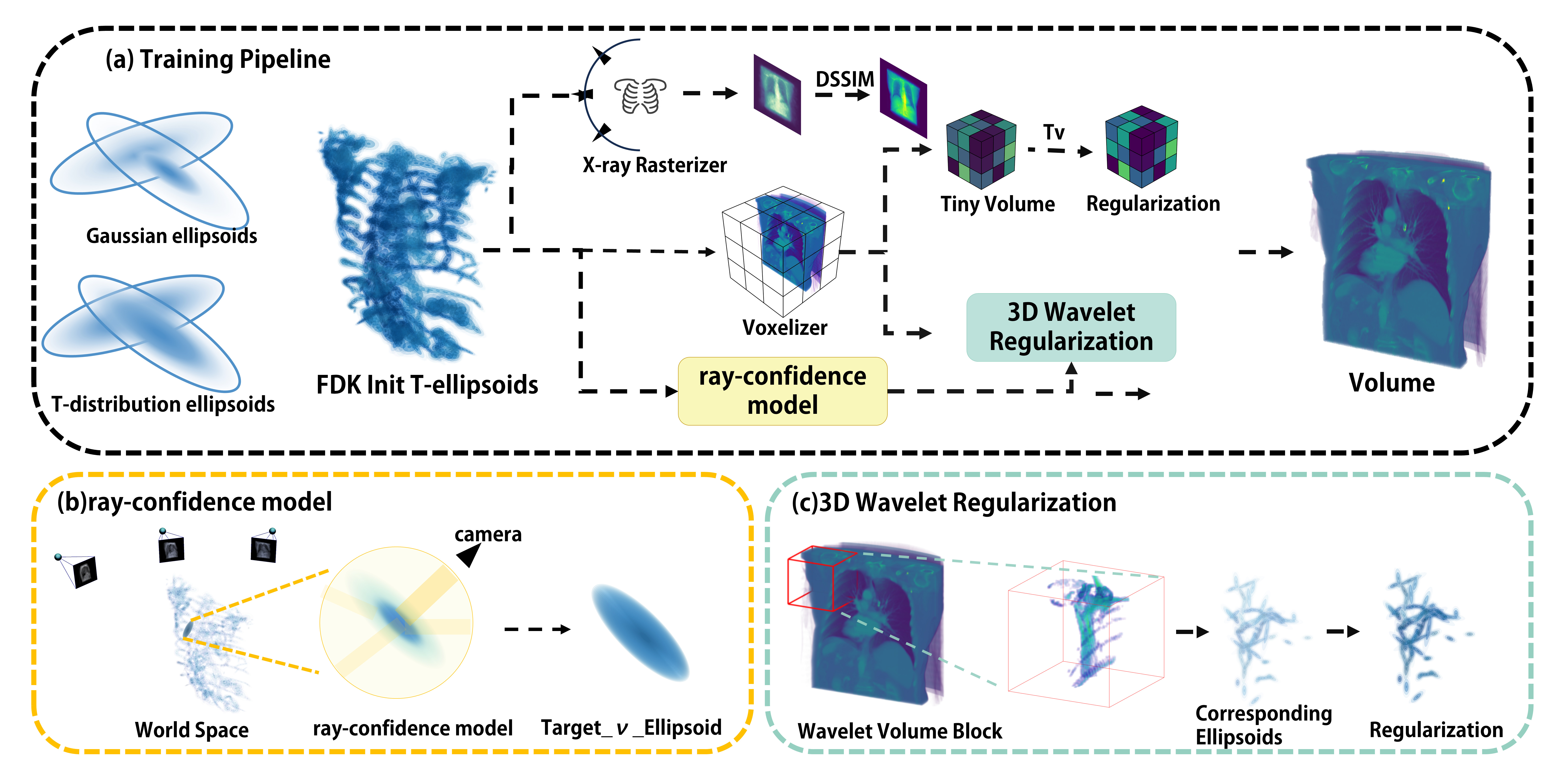}
    \caption{Training pipeline of TR-GS. (a) Overall training pipeline: TR-GS replaces Gaussian ellipsoids (thin-tail) with t-distribution ellipsoids (heavy-tail) for robustness. FDK-initialized t-ellipsoids are optimized via three parallel branches: X-ray rasterization for projection rendering, voxelization with wavelet regularization, and ray-confidence modeling. (b) Conceptual ray-confidence pipeline: geometric confidence from ray coverage is used to compute a target degree of freedom (denoted as \texttt{target\_v} in the figure, and as \(\nu_i^{\text{target}}\) in the text), which regularizes the actual \(\nu_i\) to enhance robustness for poorly-observed primitives. (c) 3D wavelet regularization: local volumetric patches are mapped to ellipsoids and enhanced in the wavelet domain to preserve anatomical details.}
    \Description{Training pipeline diagram of TR-GS showing the overall framework with three parallel optimization branches: differentiable X-ray rasterization for projection rendering, density voxelization with wavelet regularization, and ray-confidence modeling for adaptive degree-of-freedom regulation.}
    \label{fig:pipeline}
\end{figure*}

\section{Method}
\label{sec:method}
We present TR-GS, a framework for high-fidelity sparse-view CT reconstruction. It comprises three components: (1) Student's \(t\)-primitive volumetric representation, (2) ray-confidence modeling with target degree-of-freedom regularization, and (3) confidence-guided 3D wavelet volume enhancement.

\subsection{Student's t-Primitive Representation}
\label{sec:t_primitive}

As shown in Fig.~\ref{fig:pipeline}(a), our framework has three parallel branches: differentiable X-ray rasterization, density voxelization with wavelet regularization, and ray-confidence modeling. This section focuses on the t-primitive representation and its projection; the latter two are detailed in Sec.~\ref{sec:ray_confidence} and Sec.~\ref{sec:wavelet}.

We represent the object with learnable 3D Student's \(t\) kernels \(T = \{T_i\}_{i=1}^M\).

\begin{figure}[t]
    \centering
    \includegraphics[width=\columnwidth]{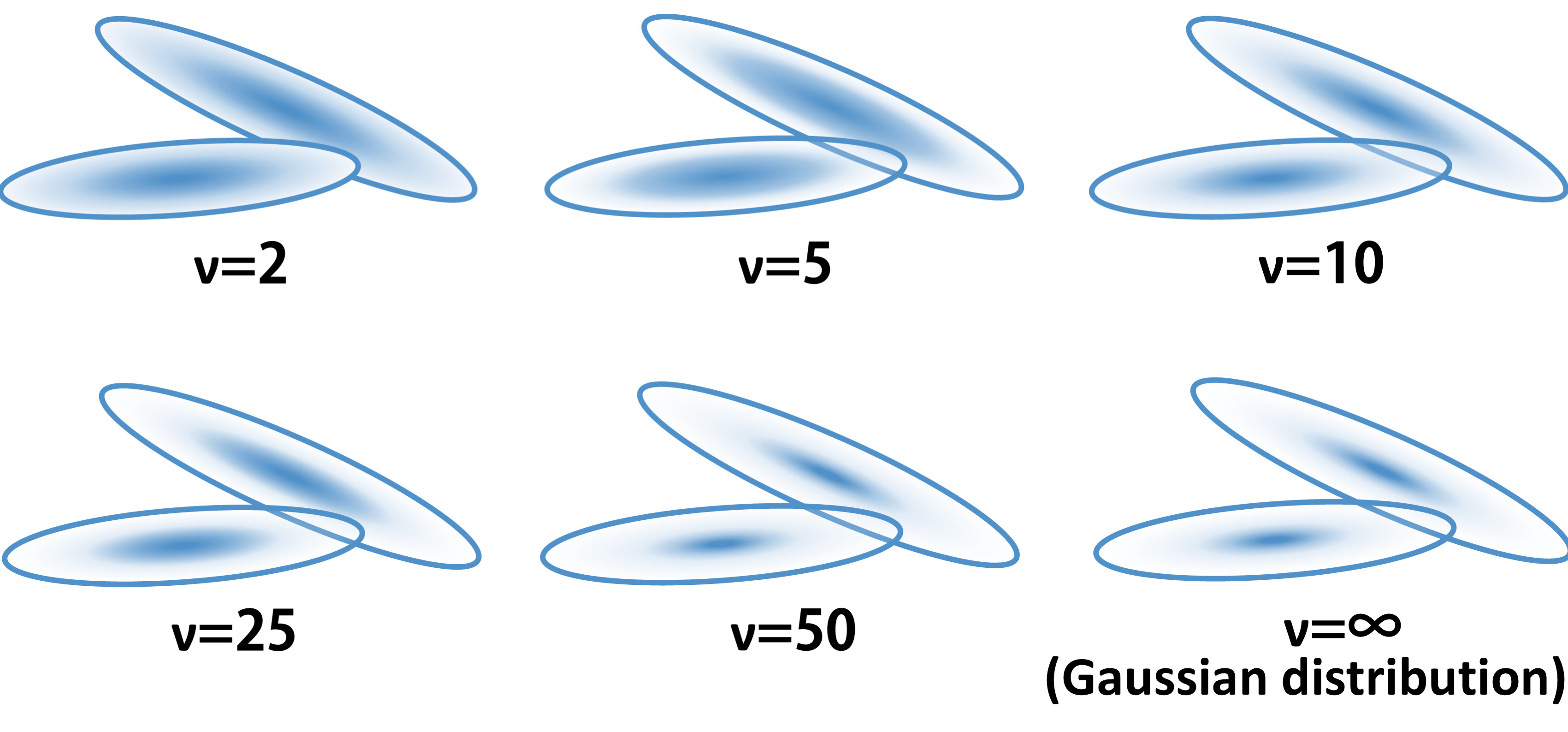}
    \caption{\(t\)-ellipsoids with varying \(\nu\). As \(\nu\) decreases, the ellipsoids exhibit heavier tails (more robust to outliers); as \(\nu \to \infty\), the \(t\) distribution converges to Gaussian.}
    \Description{Visualization of Student's t ellipsoids under different degrees of freedom. Lower values of nu produce heavier-tailed ellipsoids that are more robust to outliers; as nu approaches infinity, the t distribution converges to a Gaussian shape.}
    \label{fig:studentT}
\end{figure}

The density field is a weighted sum of anisotropic Student's \(t\) primitives:
\begin{equation}
f(\mathbf{x}) = \sum_{i=1}^{M} \rho_i \, T_i(\mathbf{x}), \label{eq:density_field}
\end{equation}
with \(\rho_i\) the density coefficient and \(T_i(\mathbf{x})\) the unnormalized kernel:
\begin{equation}
T_i(\mathbf{x}) = \left[ 1 + \frac{1}{\nu_i} (\mathbf{x}-\boldsymbol{\mu}_i)^\top \boldsymbol{\Sigma}_i^{-1} (\mathbf{x}-\boldsymbol{\mu}_i) \right]^{-\frac{\nu_i+3}{2}}. \label{eq:t_kernel_3d}
\end{equation}
Here \(\boldsymbol{\mu}_i\) is the center, \(\boldsymbol{\Sigma}_i\) the covariance, and \(\nu_i>0\) the degrees of freedom controlling tail heaviness (Fig.~\ref{fig:studentT}). The Student's t distribution is a classic robust alternative to the Gaussian in statistical modeling~\cite{murphy2012machine}. The covariance is parameterized by principal scales \((a_i,b_i,c_i)\) and rotation \(\mathbf{R}_i\):
\begin{equation}
\boldsymbol{\Sigma}_i = \mathbf{R}_i \, \mathrm{diag}(a_i^2, b_i^2, c_i^2) \, \mathbf{R}_i^\top. \label{eq:sigma_decomp}
\end{equation}

Projection along an X‑ray ray is analytic: marginalization preserves the \(t\)-form. Each ellipsoid contributes to the 2D projection according to its density distribution and viewing geometry. Let \(\tilde{\mathbf{x}}\) be ray‑aligned coordinates and \(\hat{\mathbf{x}}\) detector coordinates. The projected intensity is
\begin{align}
I_i(\hat{\mathbf{x}}) &= \rho_i \int_{-\infty}^{\infty} T_i(\tilde{\mathbf{x}}) \, dx_3 \nonumber \\
&= \rho_i \, C(\nu_i, \tilde{\boldsymbol{\Sigma}}_i)
\left[ 1 + \frac{1}{\nu_i} (\hat{\mathbf{x}} - \hat{\boldsymbol{\mu}}_i)^\top \hat{\boldsymbol{\Sigma}}_i^{-1} (\hat{\mathbf{x}} - \hat{\boldsymbol{\mu}}_i) \right]^{-\frac{\nu_i+2}{2}}, \label{eq:projected_t_corrected}
\end{align}
with the integral factor
\begin{equation}
C(\nu_i, \tilde{\boldsymbol{\Sigma}}_i)
=
\frac{\Gamma\left(\frac{\nu_i+2}{2}\right)}
{\Gamma\left(\frac{\nu_i+3}{2}\right)}
\sqrt{\nu_i \pi}\;
\frac{|\tilde{\boldsymbol{\Sigma}}_i|^{1/2}}
{|\hat{\boldsymbol{\Sigma}}_i|^{1/2}}
\label{eq:C_corrected}
\end{equation}
where \(\hat{\boldsymbol{\Sigma}}_i\) is the 2D covariance in the detector plane from the ray‑aligned 3D covariance \(\tilde{\boldsymbol{\Sigma}}_i\). Summing over primitives gives the final ray intensity.

\subsection{Ray-Confidence Modeling}
\label{sec:ray_confidence}

We compute a target degree of freedom \(\nu_i^{\text{target}}\) per primitive from ray coverage and regularize the actual \(\nu_i\) toward it; the ray‑confidence score \(c_i\in[0,1]\) is based on coverage of the unit sphere around each primitive (Fig.~\ref{fig:pipeline}(b)).

\begin{figure}[t]
    \centering
    \includegraphics[width=\columnwidth]{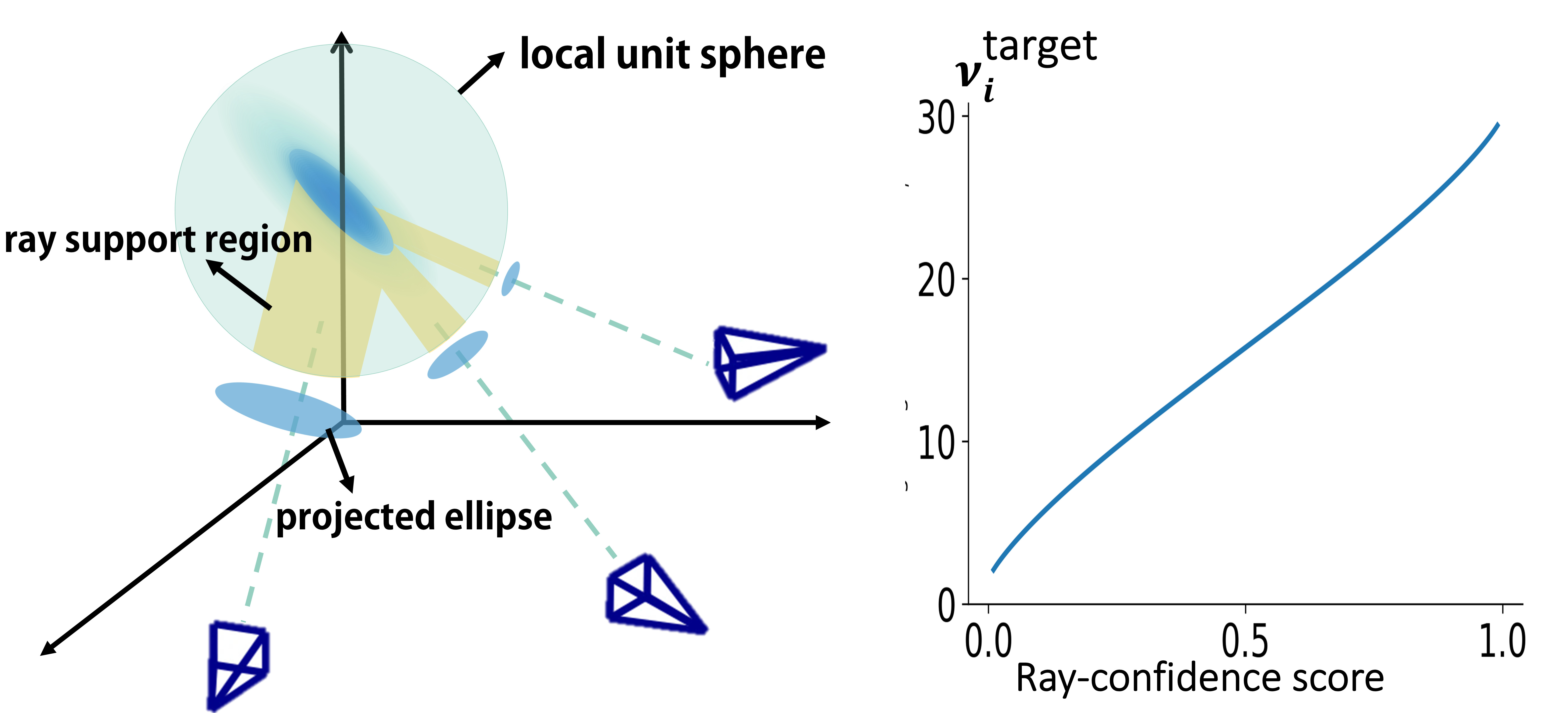}
    \caption{Illustration of ray‑confidence modeling. (Left) Geometric reliability: rays intersect the local unit sphere, forming ray support regions of varying thickness; the projected ellipse indicates the detector footprint. (Right) Mapping from confidence score \(c_i\) to target degree of freedom.}
    \Description{Illustration of ray-confidence modeling. Left: geometric reliability diagram showing rays intersecting a local unit sphere around a primitive, forming ray support regions. The projected ellipse on the detector plane is also shown. Right: mapping curve from confidence score to target degree of freedom.}
    \label{fig:Ray}
\end{figure}

For primitive \(i\) and camera \(j\), let \(\mathbf{u}_{ij}\) be the normalized direction from the primitive center to camera in local coordinates. The effective directional ratio \(e_{ij}\) is
\begin{equation}
e_{ij} = \frac{\alpha}{\sqrt{(a_i u_{ij}^{x})^2 + (b_i u_{ij}^{y})^2 + (c_i u_{ij}^{z})^2}}, \label{eq:effective_ratio}
\end{equation}
with \((a_i,b_i,c_i)\) from Eq.~\eqref{eq:sigma_decomp} and scaling factor \(\alpha\). This makes primitives of different shapes comparable.

The normalized support radius is
\begin{equation}
r_{ij} = r_{\max} \, \min\left( \frac{e_{ij}}{\kappa}, 1 \right),
\end{equation}
with empirical \(r_{\max},\kappa\). We sample \(N_s\) points \(\mathbf{p}_s\) on the unit sphere and define
\begin{equation}
c_i = \frac{1}{N_s} \sum_{s=1}^{N_s}
\mathbf{1}\left(
\exists j: \operatorname{dist}\big(\mathbf{p}_s, \text{ray segment } \mathcal{S}_{ij}\big) \le r_{ij}
\right),
\end{equation}
with \(\mathcal{S}_{ij}\) the ray segment inside the sphere. \(c_i\) measures the fraction of the sphere covered by rays.

We then compute the target degree of freedom:
\begin{equation}
\nu_i^{\text{target}} = \nu_{\min} + (\nu_{\max} - \nu_{\min}) \, \sigma\Bigg(\frac{\operatorname{logit}(c_i)}{T}\Bigg),
\end{equation}
with \(\operatorname{logit}(c_i)=\log\frac{c_i}{1-c_i}\), sigmoid \(\sigma\), and temperature \(T\). The sigmoid mapping is shown in Fig.~\ref{fig:Ray} (right). High confidence yields larger \(\nu_i^{\text{target}}\) (Gaussian‑like), low confidence yields smaller \(\nu_i^{\text{target}}\) (heavier tails).

We regularize \(\nu_i\) toward its target with a soft log‑space MSE loss:
\begin{equation}
\mathcal{L}_\nu = \frac{1}{M} \sum_{i=1}^{M} \lambda_\nu \, \omega(c_i) \, \big(\log \nu_i - \log \nu_i^{\text{target}} \big)^2,
\end{equation}
where \(\omega(c_i)\) emphasizes low‑confidence primitives.
\subsection{3D Wavelet Regularization Guided by Confidence}
\label{sec:wavelet}

\begin{figure}[t]
    \centering
    \includegraphics[width=\columnwidth]{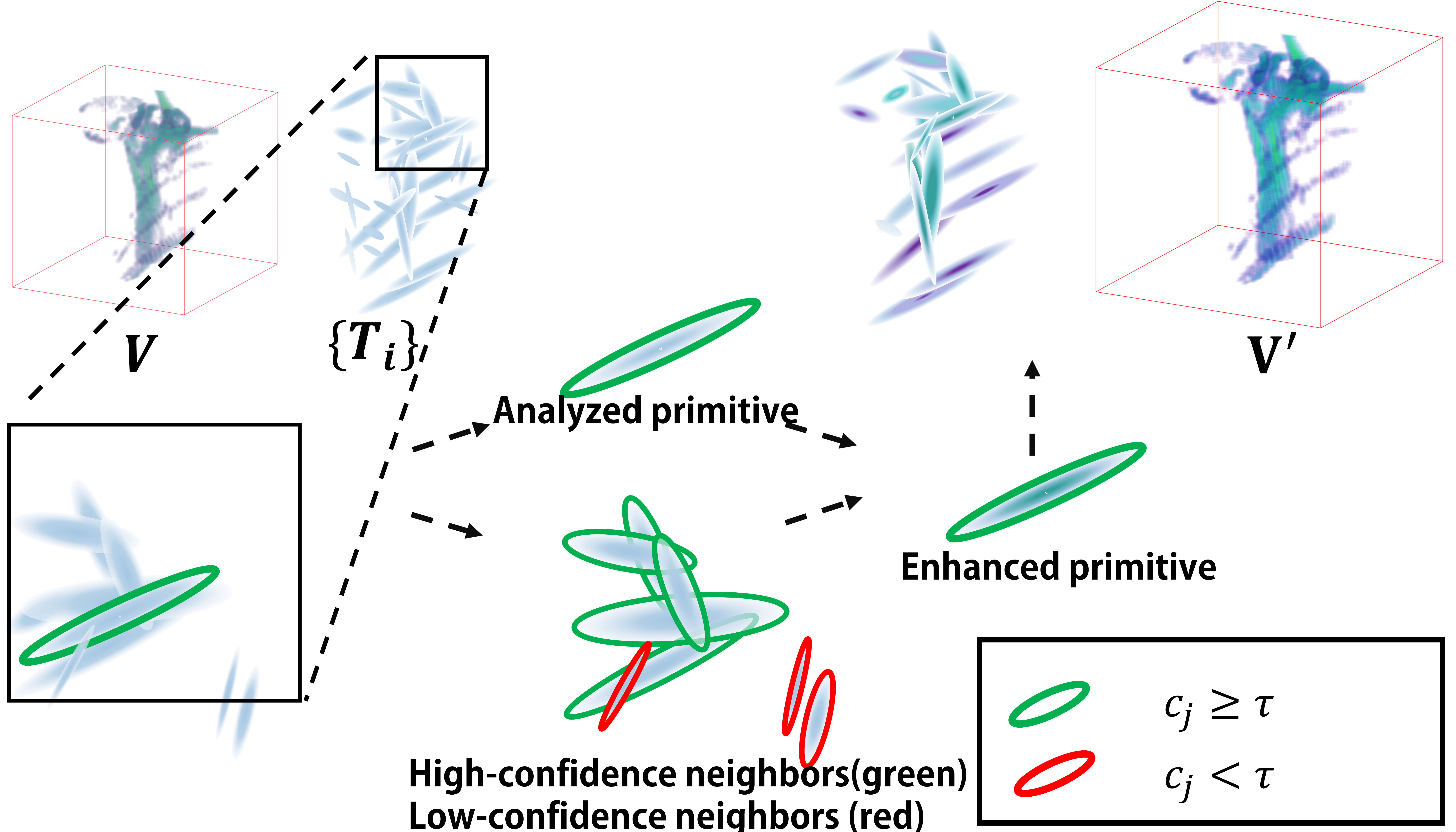}
    \caption{Confidence-guided wavelet regularization. Symbols: $\mathbf{V}$: local volume block; $\{T_i\}$: set of primitives; green/red neighbors: high‑/low‑confidence ($c_j \ge \tau$ / $c_j < \tau$); enhanced primitive (darkened) results from the wavelet loss $\mathcal{L}_{\text{wavelet}} = \lambda_{\text{supp}}S - \lambda_{\text{enh}}E$ (Eq.~\ref{eq:wavelet_loss}); final volume $\mathbf{V}'$.}
    \Description{Diagram of confidence-guided wavelet regularization. A local volumetric block is decomposed via 3D discrete wavelet transform. Neighboring primitives are partitioned into high-confidence (green) and low-confidence (red) groups based on a threshold, routing voxels to enhancement or suppression to produce the final enhanced volume.}
    \label{fig:wavelet}
\end{figure}

For smaller \(\nu\), the wider spatial support of Student's \(t\) primitives may attenuate high-frequency details. We introduce confidence-guided 3D wavelet regularization to mitigate this effect while suppressing noise.

During training, we periodically sample a local volumetric block \(\mathbf{V}\) (Fig.~\ref{fig:wavelet}) and apply a 3D discrete wavelet transform~\cite{mallat1989theory}:
\begin{equation}
\mathcal{W}(\mathbf{V}) = \{\mathbf{B}_0, \mathbf{B}_1, \dots, \mathbf{B}_{M_b-1}\},
\end{equation}
with \(\mathbf{B}_0\) low‑frequency and the rest high‑frequency. We focus on mid‑frequency subbands (indices \(1,\dots,6\) in 8‑subband decomposition), which capture structured features.

For each selected subband, we reconstruct it and compute a voxel‑wise high‑frequency strength \(g(\mathbf{p})\) (gradient magnitude). To incorporate confidence, a \(K\)‑nearest neighbor search over primitives for each voxel \(\mathbf{p}\) (visualized as \(\{T_i\}\) in Fig.~\ref{fig:wavelet}) retrieves indices \(\mathcal{N}_K(\mathbf{p})\) of the \(K\) closest Gaussians, with confidence \(c_j\). A threshold \(\tau=0.35\) partitions neighbors into high‑ and low‑confidence groups:
\begin{align}
\mathcal{N}^{\text{high}}(\mathbf{p}) &= \{j \in \mathcal{N}_K(\mathbf{p}) \mid c_j \ge \tau\},\\
\mathcal{N}^{\text{low}}(\mathbf{p}) &= \{j \in \mathcal{N}_K(\mathbf{p}) \mid c_j < \tau\},
\end{align}
colored green and red in Fig.~\ref{fig:wavelet}. The enhancement and suppression contributions per voxel are
\begin{equation}
e(\mathbf{p}) = \frac{|\mathcal{N}^{\text{high}}(\mathbf{p})|}{K}\,g(\mathbf{p}),\qquad
s(\mathbf{p}) = \frac{|\mathcal{N}^{\text{low}}(\mathbf{p})|}{K}\,g(\mathbf{p}).
\end{equation}
Averaging over all voxels gives \(E\) and \(S\). The wavelet loss is
\begin{equation}
\mathcal{L}_{\text{wavelet}} = \lambda_{\text{supp}}S - \lambda_{\text{enh}}E, \label{eq:wavelet_loss}
\end{equation}
with hyperparameters \(\lambda_{\text{supp}},\lambda_{\text{enh}}\). This encourages enhancing high-frequency details where confidence is high (negative term reduces loss) and suppressing them where confidence is low (positive term increases loss).

This regularization complements the \(t\) representation by recovering fine structures blurred by heavy tails, preserving details without amplifying noise.

\section{Experiments}

\subsection{Experimental Setup}

We evaluate TR-GS on sparse-view CT reconstruction, focusing on volumetric rendering under severely limited projection views. The experiments investigate three aspects: (1) the effect of replacing Gaussian primitives with Student's \(t\) primitives on reconstruction robustness in underconstrained settings, (2) the contribution of the ray-confidence model and wavelet regularization to reconstruction quality, and (3) the volumetric fidelity and projection consistency of TR-GS compared to representative baselines.

\paragraph{Datasets.}
We use the synthetic and real-world CT datasets from R\(^2\)-Gaussian~\cite{Zha2024R2Gaussian}, covering diverse anatomies. Main comparisons use 12, 18, and 25 uniformly sampled views over the full scanning trajectory~\cite{hanson1979detectability}. For the ablation studies, we additionally use 6- and 50-view settings to examine the components under more extreme sparse-view and relatively denser-view conditions.

\paragraph{Baselines and metrics.}
We compare TR-GS with representative methods from three categories: the analytical method FDK~\cite{FeldkampDavisKress1984}, the iterative methods ASD-POCS~\cite{SidkyPan2008} and SART~\cite{AndersenKak1984}, and neural rendering methods SAX-NeRF~\cite{Cai2024} and R\(^2\)-Gaussian~\cite{Zha2024R2Gaussian}. Following prior work, we report both volumetric and projection-domain metrics: PSNR\(_{3D}\), SSIM\(_{3D}\) evaluate the fidelity of the reconstructed 3D volume, while PSNR\(_{2D}\), SSIM\(_{2D}\) measure the consistency between rendered projections and ground-truth projections.

\paragraph{Implementation details.}
All methods use the same sparse-view settings and evaluation protocol. Projection generation and classical baseline reconstruction are performed with TIGRE~\cite{biguri2016tigre}. Unless otherwise specified, all quantitative results are computed on the same test split using the same preprocessing and normalization.

\subsection{Comparison with Baselines}

\begin{figure*}[t]
\centering
\includegraphics[width=\textwidth]{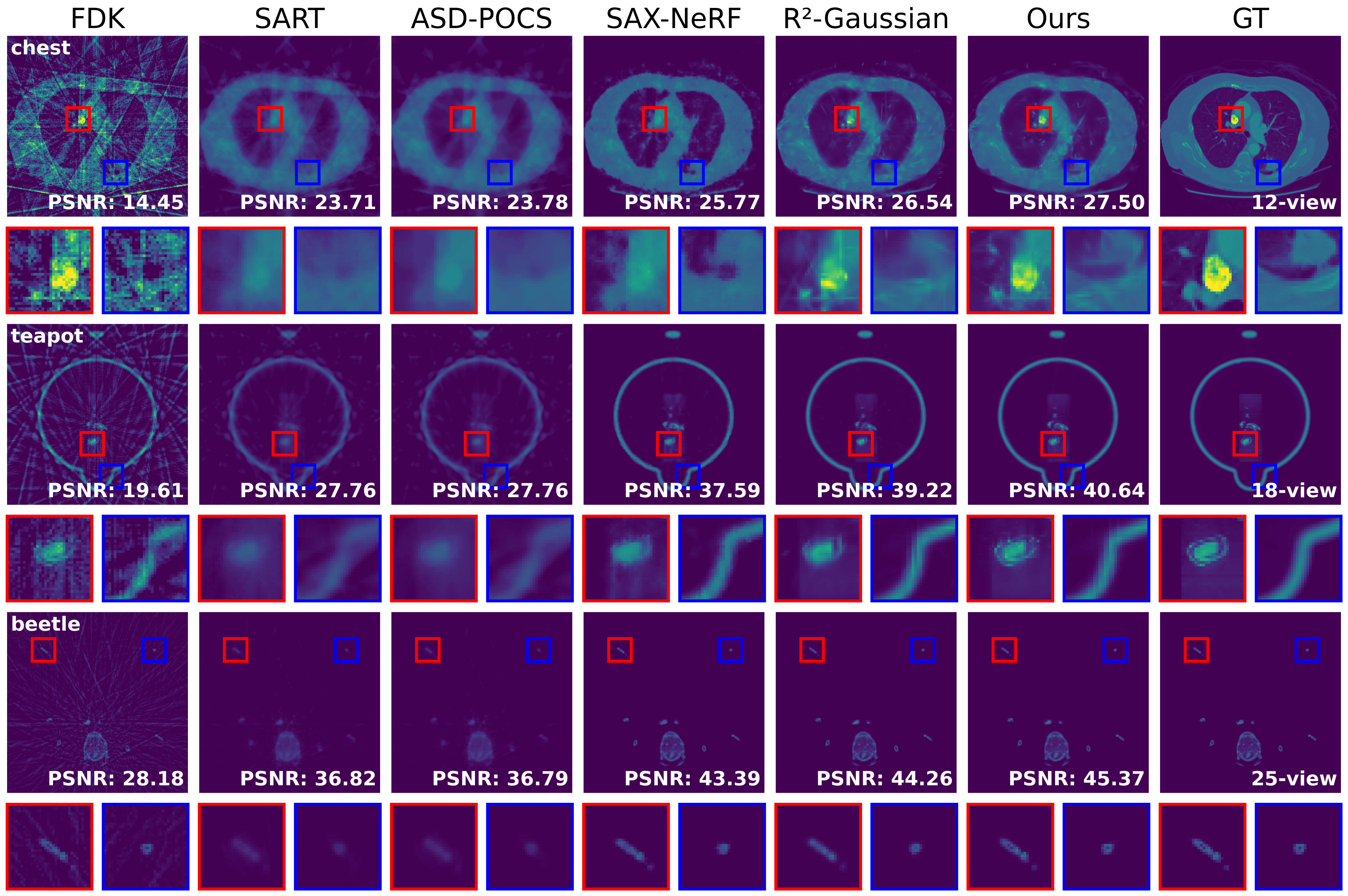}
\caption{Colorized slice examples of different methods, with PSNR (dB) shown at the bottom right of each image. In the shown examples, TR-GS recovers finer anatomical structures and exhibits fewer streak artifacts than the compared methods, particularly at 12 and 18 views.}
\Description{Colorized CT slice comparison across different reconstruction methods including FDK, ASD-POCS, SART, SAX-NeRF, R2-Gaussian, and our TR-GS method. PSNR values are shown at the bottom right of each slice image. In the shown examples, TR-GS recovers finer anatomical structures with reduced streak artifacts.}
\label{fig:colorized_slices}
\end{figure*}

We compare TR-GS against the baselines on both synthetic and real-world datasets using the 12-, 18-, and 25-view settings described above. The quantitative results are reported in Table~\ref{tab:comp_3d} (3D metrics) and Table~\ref{tab:comp_2d} (2D metrics). Figure~\ref{fig:colorized_slices} shows visual comparisons of reconstructed slices.

As shown in Table~\ref{tab:comp_3d} and Table~\ref{tab:comp_2d}, TR-GS achieves the best or competitive performance in most evaluated settings. On the synthetic dataset, it ranks first across all three view numbers in all four metrics, outperforming R$^2$-Gaussian by up to 0.38~dB in PSNR\(_{3D}\) and 0.011 in SSIM\(_{3D}\). On the real-world dataset, TR-GS achieves the best or tied-best results in all 12- and 18-view metrics, while remaining competitive at 25 views, with SSIM\(_{3D}\) (0.837) close to the best result (0.842). These results indicate that replacing Gaussian primitives with Student's \(t\) primitives improves reconstruction robustness under sparse-view acquisition.

For projection-domain metrics (Table~\ref{tab:comp_2d}), TR-GS shows advantages in most cases. On the synthetic dataset, it achieves the best PSNR\(_{2D}\) and SSIM\(_{2D}\) for all three view settings. On the real-world dataset, it performs best at 12 and 18 views and remains competitive at 25 views. These results suggest that the proposed method improves both volumetric fidelity and projection consistency during differentiable rendering.

Overall, the improvements in both 3D and 2D metrics indicate that TR-GS provides a better balance between volumetric fidelity and rendering accuracy under sparse-view acquisition.

\begin{table}[t]
\centering
\caption{Volumetric reconstruction results on the synthetic and real-world datasets under 12, 18, and 25 views. The top three results are highlighted in red, orange, and yellow, respectively.}
\Description{Comparison of volumetric PSNR and SSIM for six reconstruction methods on synthetic and real-world datasets under 12, 18, and 25 views.}
\label{tab:comp_3d}
\setlength{\tabcolsep}{3pt}
\renewcommand{\arraystretch}{1.1}
\scriptsize

\resizebox{\columnwidth}{!}{%
\begin{tabular}{lcccccc}
\toprule
\multirow{2}{*}{\textbf{Methods}}
& \multicolumn{2}{c}{\textbf{12}}
& \multicolumn{2}{c}{\textbf{18}}
& \multicolumn{2}{c}{\textbf{25}} \\
\cmidrule(lr){2-3} \cmidrule(lr){4-5} \cmidrule(lr){6-7}
& \textbf{PSNR$_{3D}$} & \textbf{SSIM$_{3D}$}
& \textbf{PSNR$_{3D}$} & \textbf{SSIM$_{3D}$}
& \textbf{PSNR$_{3D}$} & \textbf{SSIM$_{3D}$} \\
\midrule
\multicolumn{7}{c}{\textbf{Synthetic dataset}} \\
\midrule
FDK
& 16.74 & 0.224
& 18.78 & 0.296
& 20.77 & 0.371 \\
ASD-POCS
& 26.10 & 0.813
& 27.66 & 0.856
& 29.26 & 0.893 \\
SART
& 26.06 & 0.807
& 27.60 & 0.849
& 29.17 & 0.885 \\
SAX-NeRF
& \cellcolor{yellow!25}30.21 & \cellcolor{yellow!25}0.840
& \cellcolor{yellow!25}32.18 & \cellcolor{yellow!25}0.886
& \cellcolor{yellow!25}34.12 & \cellcolor{yellow!25}0.904 \\
R$^2$-Gaussian
& \cellcolor{orange!25}30.73 & \cellcolor{orange!25}0.861
& \cellcolor{orange!25}33.28 & \cellcolor{orange!25}0.900
& \cellcolor{orange!25}35.39 & \cellcolor{orange!25}0.926 \\
Ours
& \cellcolor{red!25}\textbf{31.03} & \cellcolor{red!25}\textbf{0.872}
& \cellcolor{red!25}\textbf{33.66} & \cellcolor{red!25}\textbf{0.907}
& \cellcolor{red!25}\textbf{35.77} & \cellcolor{red!25}\textbf{0.930} \\
\midrule
\multicolumn{7}{c}{\textbf{Real-world dataset}} \\
\midrule
FDK
& 18.58 & 0.236
& 20.82 & 0.319
& 20.50 & 0.328 \\
ASD-POCS
& 27.56 & 0.825
& 29.20 & 0.863
& 30.07 & 0.904 \\
SART
& 27.54 & 0.821
& 29.15 & 0.858
& 29.97 & 0.899 \\
SAX-NeRF
& \cellcolor{yellow!25}34.19 & \cellcolor{orange!25}0.875
& \cellcolor{yellow!25}36.88 & \cellcolor{orange!25}0.904
& \cellcolor{yellow!25}34.34 & \cellcolor{orange!25}0.840 \\
R$^2$-Gaussian
& \cellcolor{orange!25}34.33 & \cellcolor{yellow!25}0.867
& \cellcolor{orange!25}37.29 & \cellcolor{yellow!25}0.902
& \cellcolor{orange!25}35.45 & \cellcolor{red!25}\textbf{0.842} \\
Ours
& \cellcolor{red!25}\textbf{34.84} & \cellcolor{red!25}\textbf{0.892}
& \cellcolor{red!25}\textbf{37.81} & \cellcolor{red!25}\textbf{0.915}
& \cellcolor{red!25}\textbf{35.71} & \cellcolor{yellow!25}0.837 \\
\bottomrule
\end{tabular}%
}
\end{table}

\begin{table}[t]
\centering
\caption{Projection-domain results on the synthetic and real-world datasets under 12, 18, and 25 views. The top three results are highlighted in red, orange, and yellow, respectively.}
\Description{Comparison of projection-domain PSNR and SSIM for SAX-NeRF, R2-Gaussian, and TR-GS on synthetic and real-world datasets under 12, 18, and 25 views.}
\label{tab:comp_2d}
\setlength{\tabcolsep}{3pt}
\renewcommand{\arraystretch}{1.1}
\scriptsize

\resizebox{\columnwidth}{!}{%
\begin{tabular}{lcccccc}
\toprule
\multirow{2}{*}{\textbf{Methods}}
& \multicolumn{2}{c}{\textbf{12}}
& \multicolumn{2}{c}{\textbf{18}}
& \multicolumn{2}{c}{\textbf{25}} \\
\cmidrule(lr){2-3} \cmidrule(lr){4-5} \cmidrule(lr){6-7}
& \textbf{PSNR$_{2D}$} & \textbf{SSIM$_{2D}$}
& \textbf{PSNR$_{2D}$} & \textbf{SSIM$_{2D}$}
& \textbf{PSNR$_{2D}$} & \textbf{SSIM$_{2D}$} \\
\midrule
\multicolumn{7}{c}{\textbf{Synthetic dataset}} \\
\midrule
SAX-NeRF
& \cellcolor{yellow!25}40.08 & \cellcolor{yellow!25}0.965
& \cellcolor{yellow!25}42.46 & \cellcolor{yellow!25}0.973
& \cellcolor{yellow!25}45.36 & \cellcolor{yellow!25}0.980 \\
R$^2$-Gaussian
& \cellcolor{orange!25}40.14 & \cellcolor{orange!25}0.968
& \cellcolor{orange!25}43.90 & \cellcolor{orange!25}0.978
& \cellcolor{orange!25}46.64 & \cellcolor{orange!25}0.982 \\
Ours
& \cellcolor{red!25}\textbf{40.73} & \cellcolor{red!25}\textbf{0.972}
& \cellcolor{red!25}\textbf{44.54} & \cellcolor{red!25}\textbf{0.981}
& \cellcolor{red!25}\textbf{47.29} & \cellcolor{red!25}\textbf{0.984} \\
\midrule
\multicolumn{7}{c}{\textbf{Real-world dataset}} \\
\midrule
SAX-NeRF
& \cellcolor{orange!25}41.33 & \cellcolor{yellow!25}0.974
& \cellcolor{yellow!25}43.57 & \cellcolor{yellow!25}0.980
& \cellcolor{yellow!25}31.71 & \cellcolor{yellow!25}0.910 \\
R$^2$-Gaussian
& \cellcolor{yellow!25}40.62 & \cellcolor{orange!25}0.975
& \cellcolor{orange!25}44.79 & \cellcolor{red!25}\textbf{0.985}
& \cellcolor{red!25}\textbf{34.31} & \cellcolor{red!25}\textbf{0.948} \\
Ours
& \cellcolor{red!25}\textbf{41.55} & \cellcolor{red!25}\textbf{0.978}
& \cellcolor{red!25}\textbf{45.77} & \cellcolor{red!25}\textbf{0.985}
& \cellcolor{orange!25}33.95 & \cellcolor{red!25}\textbf{0.948} \\
\bottomrule
\end{tabular}%
}
\end{table}

\subsection{Noise Robustness Study}

We now evaluate robustness to measurement noise on the synthetic dataset under 12, 18, and 25 views, using a mixed Poisson--Gaussian noise model~\cite{tapiovaara1993snr,foi2008practical} which is widely used to characterize X-ray detector noise. Three noise levels are considered: \textit{Low} (Poisson parameter 100000, Gaussian std 10), \textit{Medium} (Poisson 50000, Gaussian std 20), and \textit{High} (Poisson 10000, Gaussian std 40), chosen to mimic realistic low-dose conditions~\cite{ma2011low}. We compare TR-GS with R\(^2\)-Gaussian to examine robustness under noisy sparse-view conditions.

Table~\ref{tab:comp_noise_3d_views_red} shows that TR-GS performs better in most configurations, with three exceptions: 12-view high-noise PSNR, 25-view medium-noise SSIM, and 25-view high-noise PSNR. Its advantage is more consistent at 12 and 18 views.

Overall, these results indicate that TR-GS improves robustness to measurement noise in most evaluated settings, where reconstruction is more challenging due to the increased ambiguity at fewer views.

\begin{table}[t]
\centering
\caption{Noise robustness on the synthetic dataset under 12, 18, and 25 views. Low, Medium, and High denote increasing mixed Poisson--Gaussian noise levels. Best results are highlighted in red.}
\Description{Comparison of volumetric PSNR and SSIM for R2-Gaussian and TR-GS under three mixed Poisson--Gaussian noise levels and three sparse-view settings.}
\label{tab:comp_noise_3d_views_red}
\setlength{\tabcolsep}{3pt}
\renewcommand{\arraystretch}{1.1}
\scriptsize

\resizebox{\columnwidth}{!}{%
\begin{tabular}{lcccccc}
\toprule
\multirow{2}{*}{\textbf{Methods}}
& \multicolumn{2}{c}{\textbf{Low Noise}}
& \multicolumn{2}{c}{\textbf{Medium Noise}}
& \multicolumn{2}{c}{\textbf{High Noise}} \\
\cmidrule(lr){2-3} \cmidrule(lr){4-5} \cmidrule(lr){6-7}
& \textbf{PSNR$_{3D}$} & \textbf{SSIM$_{3D}$}
& \textbf{PSNR$_{3D}$} & \textbf{SSIM$_{3D}$}
& \textbf{PSNR$_{3D}$} & \textbf{SSIM$_{3D}$} \\

\midrule
\multicolumn{7}{c}{\textbf{Synthetic dataset - 12-view}} \\
\midrule
R$^2$-Gaussian & 30.73 & 0.861 & 30.31 & 0.843 & \cellcolor{red!25}\textbf{28.97} & 0.820 \\
TR-GS & \cellcolor{red!25}\textbf{31.03} & \cellcolor{red!25}\textbf{0.872}
      & \cellcolor{red!25}\textbf{30.55} & \cellcolor{red!25}\textbf{0.857}
      & 28.85 & \cellcolor{red!25}\textbf{0.825} \\

\midrule
\multicolumn{7}{c}{\textbf{Synthetic dataset - 18-view}} \\
\midrule
R$^2$-Gaussian & 33.28 & 0.900 & 33.04 & 0.893 & 29.89 & 0.828 \\
TR-GS & \cellcolor{red!25}\textbf{33.66} & \cellcolor{red!25}\textbf{0.907}
      & \cellcolor{red!25}\textbf{33.30} & \cellcolor{red!25}\textbf{0.901}
      & \cellcolor{red!25}\textbf{30.05} & \cellcolor{red!25}\textbf{0.841} \\

\midrule
\multicolumn{7}{c}{\textbf{Synthetic dataset - 25-view}} \\
\midrule
R$^2$-Gaussian & 35.39 & 0.926 & 33.95 & \cellcolor{red!25}\textbf{0.914} & \cellcolor{red!25}\textbf{31.41} & 0.872 \\
TR-GS & \cellcolor{red!25}\textbf{35.77} & \cellcolor{red!25}\textbf{0.930}
      & \cellcolor{red!25}\textbf{34.00} & 0.912
      & 31.38 & \cellcolor{red!25}\textbf{0.889} \\

\bottomrule
\end{tabular}%
}
\end{table}

\subsection{Ablation Study on Core Modules}

To understand the contribution of each proposed component, we conduct ablation experiments beyond the basic Student's \(t\) primitive representation, specifically evaluating the ray-confidence model and the confidence-guided 3D wavelet regularization. Four variants are compared: (1) the base model (Student's \(t\) primitives only), (2) base + ray-confidence model, (3) base + 3D wavelet regularization, and (4) the full model combining both. All ablation results are reported on the synthetic dataset under 6, 12, 18, 25, and 50 views.

Table~\ref{tab:ablation} shows that the full model achieves the best or near-best performance in most settings. Individual variants achieve the best results for several specific metrics, indicating that the two modules are complementary but not strictly additive across all view settings and evaluation metrics.

The ray-confidence model mainly improves volumetric optimization in sparse settings where adaptive regulation of the degrees of freedom is critical, as reflected by gains in PSNR\(_{3D}\) and SSIM\(_{3D}\) (e.g., at 18 views, PSNR\(_{3D}\) increases from 33.50 to 33.61, and SSIM\(_{3D}\) from 0.891 to 0.900). In contrast, the 3D wavelet regularization contributes more to structural preservation and projection consistency, as reflected by improvements in PSNR\(_{2D}\) and SSIM\(_{2D}\) (e.g., at 50 views, PSNR\(_{2D}\) rises from 49.42 to 49.71 when adding wavelet regularization).

When combined, the two modules provide balanced performance across the evaluated metrics. The strongest configuration can vary with the view number and evaluation metric. Overall, these results suggest that while Student's \(t\) primitives provide a robust basis for sparse-view reconstruction, the ray-confidence model and wavelet regularization each contribute to reconstruction quality and structural recovery.

\begin{table}[t]
\centering
\caption{Core-module ablation on the synthetic dataset under 6, 12, 18, 25, and 50 views. Best results are highlighted in red.}
\Description{Ablation results for the Student's t base model, ray-confidence model, confidence-guided wavelet regularization, and full TR-GS model under five view settings.}
\label{tab:ablation}
\setlength{\tabcolsep}{4pt}
\renewcommand{\arraystretch}{1.1}
\scriptsize

\resizebox{\columnwidth}{!}{%
\begin{tabular}{lcccc}
\toprule
\multicolumn{5}{c}{\textbf{6-view}} \\
\midrule
\textbf{Synthetic} & \textbf{PSNR$_{3D}$} & \textbf{SSIM$_{3D}$} & \textbf{PSNR$_{2D}$} & \textbf{SSIM$_{2D}$} \\
\midrule
Base & 26.67 & 0.798 & 33.94 & 0.947 \\
Base + Ray-confidence model & \cellcolor{red!25}\textbf{26.73} & 0.803 & 33.97 & 0.946 \\
Base + 3D wavelet regularization & 26.68 & 0.798 & 33.96 & 0.947 \\
Full & 26.70 & \cellcolor{red!25}\textbf{0.805} & \cellcolor{red!25}\textbf{33.99} & \cellcolor{red!25}\textbf{0.949} \\
\bottomrule
\end{tabular}%
}

\resizebox{\columnwidth}{!}{%
\begin{tabular}{lcccc}
\toprule
\multicolumn{5}{c}{\textbf{12-view}} \\
\midrule
\textbf{Synthetic} & \textbf{PSNR$_{3D}$} & \textbf{SSIM$_{3D}$} & \textbf{PSNR$_{2D}$} & \textbf{SSIM$_{2D}$} \\
\midrule
Base & 30.95 & 0.870 & 40.71 & 0.970 \\
Base + Ray-confidence model & 30.99 & 0.871 & 40.74 & 0.973 \\
Base + 3D wavelet regularization & 30.96 & 0.868 & 40.73 & 0.972 \\
Full & \cellcolor{red!25}\textbf{31.03} & \cellcolor{red!25}\textbf{0.872} & \cellcolor{red!25}\textbf{40.78} & \cellcolor{red!25}\textbf{0.974} \\
\bottomrule
\end{tabular}%
}

\resizebox{\columnwidth}{!}{%
\begin{tabular}{lcccc}
\toprule
\multicolumn{5}{c}{\textbf{18-view}} \\
\midrule
\textbf{Synthetic} & \textbf{PSNR$_{3D}$} & \textbf{SSIM$_{3D}$} & \textbf{PSNR$_{2D}$} & \textbf{SSIM$_{2D}$} \\
\midrule
Base & 33.50 & 0.891 & 44.50 & 0.979 \\
Base + Ray-confidence model & 33.61 & 0.900 & 44.53 & 0.980 \\
Base + 3D wavelet regularization & 33.53 & 0.896 & \cellcolor{red!25}\textbf{44.58} & 0.980 \\
Full & \cellcolor{red!25}\textbf{33.66} & \cellcolor{red!25}\textbf{0.907} & 44.54 & \cellcolor{red!25}\textbf{0.981} \\
\bottomrule
\end{tabular}%
}

\resizebox{\columnwidth}{!}{%
\begin{tabular}{lcccc}
\toprule
\multicolumn{5}{c}{\textbf{25-view}} \\
\midrule
\textbf{Synthetic} & \textbf{PSNR$_{3D}$} & \textbf{SSIM$_{3D}$} & \textbf{PSNR$_{2D}$} & \textbf{SSIM$_{2D}$} \\
\midrule
Base & 35.71 & 0.928 & 47.22 & 0.983 \\
Base + Ray-confidence model & 35.74 & \cellcolor{red!25}\textbf{0.930} & 47.25 & 0.983 \\
Base + 3D wavelet regularization & 35.75 & 0.928 & 47.27 & \cellcolor{red!25}\textbf{0.984} \\
Full & \cellcolor{red!25}\textbf{35.77} & \cellcolor{red!25}\textbf{0.930} & \cellcolor{red!25}\textbf{47.29} & \cellcolor{red!25}\textbf{0.984} \\
\bottomrule
\end{tabular}%
}

\resizebox{\columnwidth}{!}{%
\begin{tabular}{lcccc}
\toprule
\multicolumn{5}{c}{\textbf{50-view}} \\
\midrule
\textbf{Synthetic} & \textbf{PSNR$_{3D}$} & \textbf{SSIM$_{3D}$} & \textbf{PSNR$_{2D}$} & \textbf{SSIM$_{2D}$} \\
\midrule
Base & 38.04 & 0.946 & 49.42 & 0.985 \\
Base + Ray-confidence model & 38.08 & 0.946 & 49.75 & 0.985 \\
Base + 3D wavelet regularization & 38.10 & \cellcolor{red!25}\textbf{0.947} & 49.71 & \cellcolor{red!25}\textbf{0.986} \\
Full & \cellcolor{red!25}\textbf{38.17} & \cellcolor{red!25}\textbf{0.947} & \cellcolor{red!25}\textbf{49.77} & 0.985 \\
\bottomrule
\end{tabular}%
}

\end{table}

\subsection{Ablation Study on Optimizer and Density Control}

In addition to the proposed algorithmic modules, TR-GS adopts specific engineering choices for training. TR-GS uses Stochastic Gradient Hamiltonian Monte Carlo (SGHMC) for position parameters \(\mu\) and Adam for the remaining parameters, with an Adding \& Recycling (AR) densification mechanism following~\cite{zhu20253d}. R\(^2\)-Gaussian uses Adaptive Density Control (ADC) and Adam by default. These engineering choices are not claimed as primary innovations. We evaluate all four combinations of density control (ADC vs.\ AR) and optimizer (Adam vs.\ SGHMC) for both methods under the 18-view low-noise setting to assess whether the performance difference stems from these engineering choices.

Table~\ref{tab:comp_density_opt} reports the results. For R\(^2\)-Gaussian, the default configuration (ADC+Adam) achieves the highest PSNR\(_{3D}\) (33.28~dB) and SSIM\(_{3D}\) (0.900), confirming that ADC and Adam are well suited for Gaussian primitives. For TR-GS, the best performance is obtained with AR+SGHMC (33.66~dB, 0.907), which is our default configuration. Within TR-GS, AR outperforms ADC when paired with the same optimizer (e.g., AR+SGHMC vs.\ ADC+SGHMC: 33.66 vs.\ 33.57), and SGHMC yields higher results than Adam. Under matched ADC+Adam, TR-GS retains a small numerical advantage (33.37~dB vs.\ 33.28~dB), while applying AR+SGHMC to R\(^2\)-Gaussian does not close the gap (32.47~dB vs.\ 33.66~dB).

These results provide two observations. First, each method performs best with its own default configuration, and the Student's \(t\)-based model retains an advantage even under matched engineering choices. Second, the optimizer and densification mechanism alone cannot account for the observed performance difference, suggesting that the primitive representation contributes to the improvement. We therefore adopt AR+SGHMC as it is empirically validated for the Student's \(t\) setting, while noting that these are implementation choices rather than core contributions.

As an additional analysis, we examined whether the learned degrees of freedom \(\nu_i\) align with the ray-confidence scores \(c_i\), to verify that the confidence model meaningfully conditions the primitive distribution. Table~\ref{tab:sensitivity_T} reports the Spearman rank correlation and reconstruction metrics under different temperature settings on the synthetic 12-view setting. While reducing \(T\) to 0.1 strengthens the \(\nu\)--\(c\) alignment (\(\rho=0.87\)), it does not yield the best reconstruction quality. The default \(T=0.25\) achieves the best PSNR and SSIM with a moderate correlation (\(\rho=0.66\)), indicating that the learned \(\nu\) values spatially track ray-confidence scores and that a balanced rather than maximal alignment is preferable.

\begin{table}[t]
\centering
\caption{Sensitivity of the ray-confidence temperature \(T\) on the synthetic 12-view setting.}
\Description{Sensitivity analysis of the temperature parameter in the ray-confidence model, showing Spearman rank correlation between learned degrees of freedom and confidence scores, along with reconstruction metrics.}
\label{tab:sensitivity_T}
\setlength{\tabcolsep}{4pt}
\renewcommand{\arraystretch}{1.1}
\scriptsize
\resizebox{\columnwidth}{!}{%
\begin{tabular}{lccccc}
\toprule
\(T\) & Spearman \(\rho\) & PSNR\(_{3D}\) \(\uparrow\) & SSIM\(_{3D}\) \(\uparrow\) & PSNR\(_{2D}\) \(\uparrow\) & SSIM\(_{2D}\) \(\uparrow\) \\
\midrule
0.10 & 0.87 & 30.90 & 0.870 & 40.79 & 0.972 \\
0.25 & 0.66 & 31.05 & 0.874 & 40.79 & 0.974 \\
0.50 & 0.56 & 30.96 & 0.867 & 40.75 & 0.973 \\
\bottomrule
\end{tabular}%
}
\end{table}

\begin{table}[t]
\centering
\caption{Optimizer and density-control ablation under the 18-view low-noise setting. Best results within each method are highlighted in red.}
\Description{Comparison of Adam and SGHMC optimizers with ADC and AR density-control strategies for R2-Gaussian and TR-GS.}
\label{tab:comp_density_opt}
\setlength{\tabcolsep}{2pt}
\renewcommand{\arraystretch}{1.1}
\resizebox{\columnwidth}{!}{%
\begin{tabular}{l c c c c}
\toprule
\textbf{Method} & \textbf{Density Control} & \textbf{Optimizer} & \textbf{PSNR$_{3D}$ (dB)} & \textbf{SSIM$_{3D}$} \\
\midrule
\multirow{4}{*}{R$^2$-Gaussian}
& ADC & Adam      & 33.28 & 0.900 \\
& AR  & Adam      & 32.44 & 0.891 \\
& AR  & SGHMC     & 32.47 & 0.892 \\
& ADC & SGHMC     & 32.41 & 0.891 \\
\midrule
\multirow{4}{*}{TR-GS (Ours)}
& AR  & SGHMC     & \cellcolor{red!25}\textbf{33.66} & \cellcolor{red!25}\textbf{0.907} \\
& AR  & Adam      & 33.45 & 0.904 \\
& ADC & Adam      & 33.37 & 0.903 \\
& ADC & SGHMC     & 33.57 & 0.905 \\
\bottomrule
\end{tabular}%
}
\end{table}

\section{Discussion and Conclusion}

We presented \textbf{TR-GS}, a Gaussian-splatting-based framework that replaces conventional Gaussian primitives with Student's \(t\)-distributions to improve robustness to outliers and incomplete observations in sparse-view CT reconstruction. The key components include a ray-confidence model that adaptively regulates the degrees of freedom according to local ray observability, and a confidence-guided 3D wavelet regularization that balances high-frequency detail preservation with noise suppression. Across the evaluated datasets and view settings, TR-GS improves over representative baselines in most cases and remains competitive in the remaining cases, with particularly strong results in highly sparse settings such as 12 and 18 views. The resulting volumetric representations may support medical multimedia applications including XR-based visualization and interactive clinical rendering. Future work includes extending the evaluation to limited-angle acquisition and broader anatomical regions, and investigating learning-based confidence estimation strategies.

In summary, we presented a reconstruction framework that replaces Gaussian primitives with projectable Student's \(t\)-distributions and couples their degrees of freedom with local ray observability, supported by confidence-guided wavelet regularization.

\begin{acks}
We sincerely thank the anonymous reviewers for their valuable feedback
and suggestions. We also thank the members of our laboratory for their
helpful discussions and support during this work. This work was supported by the National Key Research and Development Program of China (2024YFF0507801), the National Natural Science Foundation of China (62505189) and the Scientific Foundation for Youth Scholars of Shenzhen University (868-000001033229).
\end{acks}

\bibliographystyle{ACM-Reference-Format}
\bibliography{sample-base}

@article{radon1986determination,
  title={On the determination of functions from their integral values along certain manifolds},
  author={Radon, Johann},
  journal={IEEE transactions on medical imaging},
  volume={5},
  number={4},
  pages={170--176},
  year={1986},
  publisher={IEEE}
}

@inproceedings{cai2024radiative,
  author    = {Cai, Yuanhao and Liang, Yixun and Wang, Jiahao and Wang, Angtian and Zhang, Yulun and Yang, Xiaokang and Zhou, Zongwei and Yuille, Alan},
  title     = {Radiative {G}aussian Splatting for Efficient {X}-Ray Novel View Synthesis},
  booktitle = {Computer Vision --- ECCV 2024},
  pages     = {283--299},
  year      = {2025},
  publisher = {Springer}
}

@inproceedings{zhu20253d,
  title={3d student splatting and scooping},
  author={Zhu, Jialin and Yue, Jiangbei and He, Feixiang and Wang, He},
  booktitle={Proceedings of the Computer Vision and Pattern Recognition Conference},
  pages={21045--21054},
  year={2025}
}

@article{Hounsfield1980,
  author  = {Hounsfield, G. N.},
  title   = {Computed medical imaging},
  journal = {Science},
  volume  = {210},
  number  = {4465},
  pages   = {22--28},
  year    = {1980}
}

@article{Cormack1963,
  author  = {Cormack, A. M.},
  title   = {Representation of a function by its line integrals, with some radiological applications},
  journal = {Journal of Applied Physics},
  volume  = {34},
  year    = {1963},
  pages   = {2722--2727}
}

@book{KakSlaney2001,
  author    = {Kak, Avinash C. and Slaney, Malcolm},
  title     = {Principles of Computerized Tomographic Imaging},
  publisher = {SIAM},
  year      = {2001},
  isbn      = {9780898714991}
}

@article{DeChiffre2014,
  author  = {De Chiffre, Leonardo and Carmignato, Simone and Kruth, J.-P. and Schmitt, Robert and Weckenmann, Albert},
  title   = {Industrial applications of computed tomography},
  journal = {CIRP Annals},
  volume  = {63},
  number  = {2},
  pages   = {655--677},
  year    = {2014}
}

@article{Mostafapour2024,
  author  = {Mostafapour, Samaneh and Greuter, Marcel and van Snick, Johannes H. and Brouwers, Adrienne H. and Dierckx, Rudi A. J. O. and van Sluis, Joyce and Lammertsma, Adriaan A. and Tsoumpas, Charalampos},
  title   = {Ultra-low dose {CT} scanning for {PET/CT}},
  journal = {Medical Physics},
  volume  = {51},
  number  = {1},
  pages   = {139--155},
  year    = {2024}
}

@article{Koch2024,
  author  = {Koch, A. and Gruber-Rouh, Tatjana and Zangos, Stephan and Eichler, Katrin and Vogl, T. and Basten, L.},
  title   = {Radiation protection in {CT}-guided interventions: does real-time dose visualisation lead to a reduction in radiation dose to participating radiologists? {A} single-centre evaluation},
  journal = {Clinical Radiology},
  volume  = {79},
  number  = {6},
  pages   = {e785--e790},
  year    = {2024}
}

@article{DavidOlawade2025,
  author  = {Clement David-Olawade, Aanuoluwapo and Olawade, David B. and Vanderbloemen, Laura and Rotifa, Oluwayomi B. and Fidelis, Sandra Chinaza and Egbon, Eghosasere and Akpan, Akwaowo Owoidighe and Adeleke, Sola and Ghose, Aruni and Boussios, Stergios},
  title   = {{AI}-driven advances in low-dose imaging and enhancement—a review},
  journal = {Diagnostics},
  volume  = {15},
  number  = {6},
  pages   = {689},
  year    = {2025}
}

@article{Shieh2019,
  author    = {Shieh, Chun-Chien and Gonzalez, Yesenia and Li, Bin and Jia, Xun and Rit, Simon and Mory, Cyril and Riblett, Matthew and Hugo, Geoffrey and Zhang, Yawei and Jiang, Zhuoran and others},
  title     = {{SPARE}: Sparse-view reconstruction challenge for {4D} cone-beam {CT} from a 1-min scan},
  journal   = {Medical Physics},
  volume    = {46},
  number    = {9},
  pages     = {3799--3811},
  year      = {2019}
}

@article{Keall2004,
  author    = {Keall, Paul},
  title     = {4-Dimensional Computed Tomography Imaging and Treatment Planning},
  journal   = {Seminars in Radiation Oncology},
  volume    = {14},
  number    = {1},
  pages     = {81--90},
  year      = {2004},
  doi       = {10.1053/j.semradonc.2003.10.006}
}

@article{RamachandranLakshminarayanan1971,
  author  = {Ramachandran, G. N. and Lakshminarayanan, A. V.},
  title   = {Three-dimensional reconstruction from radiographs and electron micrographs: application of convolutions instead of Fourier transforms},
  journal = {Proceedings of the National Academy of Sciences},
  volume  = {68},
  number  = {9},
  pages   = {2236--2240},
  year    = {1971}
}

@article{FeldkampDavisKress1984,
  author  = {Feldkamp, Lee A. and Davis, Lloyd C. and Kress, James W.},
  title   = {Practical cone-beam algorithm},
  journal = {Journal of the Optical Society of America A},
  volume  = {1},
  number  = {6},
  pages   = {612--619},
  year    = {1984}
}

@article{AndersenKak1984,
  author  = {Andersen, Anders H. and Kak, Avinash C.},
  title   = {Simultaneous algebraic reconstruction technique (SART): a superior implementation of the ART algorithm},
  journal = {Ultrasonic Imaging},
  volume  = {6},
  number  = {1},
  pages   = {81--94},
  year    = {1984}
}

@article{SidkyPan2008,
  author  = {Sidky, Emil Y. and Pan, Xiaochuan},
  title   = {Image reconstruction in circular cone-beam computed tomography by constrained, total-variation minimization},
  journal = {Physics in Medicine \& Biology},
  volume  = {53},
  number  = {17},
  pages   = {4777},
  year    = {2008}
}

@inproceedings{Ying2019,
  author    = {Ying, Xingde and Guo, Heng and Ma, Kai and Wu, Jian and Weng, Zhengxin and Zheng, Yefeng},
  title     = {X2CT-GAN: Reconstructing CT from biplanar X-rays with generative adversarial networks},
  booktitle = {Proceedings of the IEEE/CVF Conference on Computer Vision and Pattern Recognition},
  pages     = {10619--10628},
  year      = {2019}
}

@article{Lantz2024,
  author  = {Lantz, Megan and Sidky, Emil Y. and Reiser, Ingrid S. and Pan, Xiaochuan and Ongie, Gregory},
  title   = {Enhancing signal detectability in learning-based CT reconstruction with a model-observer inspired loss function},
  journal = {arXiv preprint arXiv:2402.10010},
  year    = {2024}
}

@inproceedings{Chung2023,
  author    = {Chung, Hyungjin and Ryu, Dohoon and McCann, Michael T. and Klasky, Marc L. and Ye, Jong Chul},
  title     = {Solving 3D inverse problems using pre-trained 2D diffusion models},
  booktitle = {Proceedings of the IEEE/CVF Conference on Computer Vision and Pattern Recognition},
  pages     = {22542--22551},
  year      = {2023}
}

@article{Liu2020,
  author  = {Liu, Zhengchun and Bicer, Tekin and Kettimuthu, Rajkumar and Gursoy, Doga and De Carlo, Francesco and Foster, Ian},
  title   = {TomoGAN: low-dose synchrotron X-ray tomography with generative adversarial networks},
  journal = {JOSA A},
  volume  = {37},
  number  = {3},
  pages   = {422--434},
  year    = {2020}
}

@inproceedings{Liu2023,
  author    = {Liu, Jiaming and Anirudh, Rushil and Thiagarajan, Jayaraman J. and He, Stewart and Mohan, K. Aditya and Kamilov, Ulugbek S. and Kim, Hyojin},
  title     = {{DOLCE}: A Model-Based Probabilistic Diffusion Framework for Limited-Angle {CT} Reconstruction},
  booktitle = {Proceedings of the IEEE/CVF International Conference on Computer Vision (ICCV)},
  pages     = {10464--10474},
  year      = {2023}
}

@inproceedings{Anirudh2018,
  author    = {Anirudh, Rushil and Kim, Hyojin and Thiagarajan, Jayaraman J. and Mohan, K. Aditya and Champley, Kyle and Bremer, Timo},
  title     = {Lose the views: Limited angle CT reconstruction via implicit sinogram completion},
  booktitle = {Proceedings of the IEEE Conference on Computer Vision and Pattern Recognition},
  pages     = {6343--6352},
  year      = {2018}
}

@inproceedings{Zang2021,
  author    = {Zang, Guangming and Idoughi, Ramzi and Li, Rui and Wonka, Peter and Heidrich, Wolfgang},
  title     = {IntraTomo: Self-supervised learning-based tomography via sinogram synthesis and prediction},
  booktitle = {Proceedings of the IEEE/CVF International Conference on Computer Vision},
  pages     = {1960--1970},
  year      = {2021}
}

@inproceedings{Zha2022,
  author    = {Zha, Ruyi and Zhang, Yanhao and Li, Hongdong},
  title     = {{NAF}: Neural Attenuation Fields for Sparse-View {CBCT} Reconstruction},
  booktitle = {International Conference on Medical Image Computing and Computer-Assisted Intervention (MICCAI)},
  pages     = {442--452},
  year      = {2022},
  publisher = {Springer}
}

@inproceedings{Cai2024,
  author    = {Cai, Yuanhao and Wang, Jiahao and Yuille, Alan and Zhou, Zongwei and Wang, Angtian},
  title     = {Structure-aware sparse-view X-ray 3D reconstruction},
  booktitle = {Proceedings of the IEEE/CVF Conference on Computer Vision and Pattern Recognition},
  pages     = {11174--11183},
  year      = {2024}
}

@inproceedings{Lin2023,
  author    = {Lin, Yiqun and Luo, Zhongjin and Zhao, Wei and Li, Xiaomeng},
  title     = {Learning deep intensity field for extremely sparse-view CBCT reconstruction},
  booktitle = {MICCAI},
  pages     = {13--23},
  year      = {2023}
}

@article{Kerbl2023,
  author    = {Kerbl, Bernhard and Kopanas, Georgios and Leimk{\"u}hler, Thomas and Drettakis, George},
  title     = {3{D} {G}aussian Splatting for Real-Time Radiance Field Rendering},
  journal   = {{ACM} Trans. Graph.},
  volume    = {42},
  number    = {4},
  pages     = {139:1--139:14},
  year      = {2023},
  doi       = {10.1145/3592433}
}

@article{Nikolakakis2024,
  author  = {Nikolakakis, Emmanouil and Gupta, Utkarsh and Vengosh, Jonathan and Bui, Justin and Marinescu, Razvan},
  title   = {GASPCT: Gaussian splatting for novel CT projection view synthesis},
  journal = {arXiv preprint arXiv:2404.03126},
  year    = {2024}
}

@article{Lin2023SparseView,
  author  = {Lin, Yiqun and Luo, Zhongjin and Zhao, Wei and Li, Xiaomeng},
  title   = {Sparse-view CT reconstruction with 3D Gaussian volumetric representation},
  journal = {arXiv preprint arXiv:2312.15676},
  year    = {2023}
}

@article{Mildenhall2020,
  author    = {Mildenhall, Ben and Srinivasan, Pratul P. and Tancik, Matthew and Barron, Jonathan T. and Ramamoorthi, Ravi and Ng, Ren},
  title     = {{NeRF}: Representing Scenes as Neural Radiance Fields for View Synthesis},
  journal   = {Commun. {ACM}},
  volume    = {65},
  number    = {1},
  pages     = {99--106},
  year      = {2022},
  doi       = {10.1145/3503250}
}

@article{Ruckert2022,
  author  = {Rückert, Darius and Wang, Yuanhao and Li, Rui and Idoughi, Ramzi and Heidrich, Wolfgang},
  title   = {NeAT: Neural adaptive tomography},
  journal = {ACM Transactions on Graphics},
  volume  = {41},
  number  = {4},
  year    = {2022}
}

@article{Muller2022,
  author    = {M{\"u}ller, Thomas and Evans, Alex and Schied, Christoph and Keller, Alexander},
  title     = {Instant Neural Graphics Primitives with a Multiresolution Hash Encoding},
  journal   = {ACM Trans. Graph.},
  volume    = {41},
  number    = {4},
  pages     = {1--15},
  year      = {2022}
}

@inproceedings{GuedonLepetit2024,
  author    = {Gu{\'e}don, Antoine and Lepetit, Vincent},
  title     = {{SuGaR}: Surface-Aligned {G}aussian Splatting for Efficient {3D} Mesh Reconstruction and High-Quality Mesh Rendering},
  booktitle = {Proceedings of the IEEE/CVF Conference on Computer Vision and Pattern Recognition (CVPR)},
  pages     = {5354--5363},
  year      = {2024}
}

@inproceedings{Lu2024,
  author    = {Lu, Tao and Yu, Mulin and Xu, Linning and Xiangli, Yuanbo and Wang, Limin and Lin, Dahua and Dai, Bo},
  title     = {{Scaffold-GS}: Structured {3D} {G}aussians for View-Adaptive Rendering},
  booktitle = {Proceedings of the IEEE/CVF Conference on Computer Vision and Pattern Recognition (CVPR)},
  pages     = {20654--20664},
  year      = {2024}
}

@inproceedings{Tang2024,
  author    = {Tang, Jiaxiang and Ren, Jiawei and Zhou, Hang and Liu, Ziwei and Zeng, Gang},
  title     = {{DreamGaussian}: Generative {G}aussian Splatting for Efficient {3D} Content Creation},
  booktitle = {International Conference on Learning Representations (ICLR)},
  pages     = {33879--33896},
  year      = {2024}
}

@inproceedings{Matsuki2024,
  author    = {Matsuki, Hidenobu and Murai, Riku and Kelly, Paul H. J. and Davison, Andrew J.},
  title     = {Gaussian Splatting SLAM},
  booktitle = {Proceedings of the IEEE/CVF Conference on Computer Vision and Pattern Recognition},
  pages     = {18039--18048},
  year      = {2024}
}

@article{Zha2024R2Gaussian,
  author  = {Zha, Ruyi and Lin, Tao Jun and Cai, Yuanhao and Cao, Jiwen and Zhang, Yanhao and Li, Hongdong},
  title   = {R$^2$-Gaussian: Rectifying radiative Gaussian splatting for tomographic reconstruction},
  journal = {arXiv preprint arXiv:2405.20693},
  year    = {2024}
}

@article{brenner2007computed,
  title={Computed tomography—an increasing source of radiation exposure},
  author={Brenner, David J and Hall, Eric J},
  journal={New England journal of medicine},
  volume={357},
  number={22},
  pages={2277--2284},
  year={2007},
  publisher={Mass Medical Soc}
}

@article{mccollough2009strategies,
  title={Strategies for reducing radiation dose in CT},
  author={McCollough, Cynthia H and Primak, Andrew N and Braun, Natalie and Kofler, James and Yu, Lifeng and Christner, Jodie},
  journal={Radiologic Clinics of North America},
  volume={47},
  number={1},
  pages={27},
  year={2009}
}

@article{mallat1989theory,
  title={A theory for multiresolution signal decomposition: the wavelet representation},
  author={Mallat, Stephane G},
  journal={IEEE transactions on pattern analysis and machine intelligence},
  volume={11},
  number={7},
  pages={674--693},
  year={1989},
  publisher={IEEE}
}

@article{barsom2016systematic,
  title={Systematic review on the effectiveness of augmented reality applications in medical training},
  author={Barsom, Esther Z and Graafland, Maurits and Schijven, Marlies P},
  journal={Surgical endoscopy},
  volume={30},
  number={10},
  pages={4174--4183},
  year={2016},
  publisher={Springer}
}

@book{murphy2012machine,
  title={Machine learning: a probabilistic perspective},
  author={Murphy, Kevin P},
  year={2012},
  publisher={MIT press}
}

@article{sidky2006accurate,
  title={Accurate image reconstruction from few-views and limited-angle data in divergent-beam CT},
  author={Sidky, Emil Y and Kao, Chien-Min and Pan, Xiaochuan},
  journal={Journal of X-ray Science and Technology},
  volume={14},
  number={2},
  pages={119--139},
  year={2006},
  publisher={SAGE Publications Sage UK: London, England}
}

@article{wolterink2017generative,
  title={Generative adversarial networks for noise reduction in low-dose CT},
  author={Wolterink, Jelmer M and Leiner, Tim and Viergever, Max A and I{\v{s}}gum, Ivana},
  journal={IEEE transactions on medical imaging},
  volume={36},
  number={12},
  pages={2536--2545},
  year={2017},
  publisher={IEEE}
}

@article{biguri2016tigre,
  title={TIGRE: a MATLAB-GPU toolbox for CBCT image reconstruction},
  author={Biguri, Ander and Dosanjh, Manjit and Hancock, Steven and Soleimani, Manuchehr},
  journal={Biomedical Physics \& Engineering Express},
  volume={2},
  number={5},
  pages={055010},
  year={2016},
  publisher={IOP Publishing}
}

@article{tapiovaara1993snr,
  title={SNR and noise measurements for medical imaging: I. A practical approach based on statistical decision theory},
  author={Tapiovaara, MJ and Wagner, RF},
  journal={Physics in Medicine \& Biology},
  volume={38},
  number={1},
  pages={71--92},
  year={1993}
}

@article{ma2011low,
  title={Low-dose computed tomography image restoration using previous normal-dose scan},
  author={Ma, Jianhua and Huang, Jing and Feng, Qianjin and Zhang, Hua and Lu, Hongbing and Liang, Zhengrong and Chen, Wufan},
  journal={Medical physics},
  volume={38},
  number={10},
  pages={5713--5731},
  year={2011},
  publisher={Wiley Online Library}
}

@article{foi2008practical,
  title={Practical Poissonian-Gaussian noise modeling and fitting for single-image raw-data},
  author={Foi, Alessandro and Trimeche, Mejdi and Katkovnik, Vladimir and Egiazarian, Karen},
  journal={IEEE transactions on image processing},
  volume={17},
  number={10},
  pages={1737--1754},
  year={2008},
  publisher={IEEE}
}

@article{hanson1979detectability,
  title={Detectability in computed tomographic images},
  author={Hanson, Kenneth M},
  journal={Medical Physics},
  volume={6},
  number={5},
  pages={441--451},
  year={1979},
  publisher={Wiley Online Library}
}

\clearpage
\begin{strip}
\centering
{\LARGE \bfseries Supplementary Material\par}
\vspace{1em}
\end{strip}

\appendix

\section{Detailed Derivation of the Projected Student's \(t\)-Primitive}
\label{app:projected_t_derivation}

This appendix provides the intermediate derivation steps omitted in the main paper. In particular, we expand the closed-form X-ray projection of a single 3D Student's \(t\)-primitive, as well as the corresponding amplitude factor. Throughout this appendix, we follow exactly the notation in the main text and only make the ray-aligned coordinates more explicit for derivation purposes.

\subsection{Ray-aligned coordinates}

In the main text, the \(i\)-th 3D Student's \(t\)-primitive is defined as
\begin{equation}
T_i(\mathbf{x}) =
\left[
1 + \frac{1}{\nu_i}
(\mathbf{x}-\boldsymbol{\mu}_i)^\top
\boldsymbol{\Sigma}_i^{-1}
(\mathbf{x}-\boldsymbol{\mu}_i)
\right]^{-\frac{\nu_i+3}{2}}.
\end{equation}

To evaluate its contribution along an X-ray ray, we rewrite it in a local ray-aligned coordinate system. Let
\begin{equation}
\tilde{\mathbf{x}}=(x_1,x_2,x_3)
\end{equation}
denote the ray-aligned 3D coordinates, where \(x_3\) is aligned with the ray direction and \((x_1,x_2)\) span the detector plane. We denote the corresponding detector-plane coordinate by
\begin{equation}
\hat{\mathbf{x}}=(x_1,x_2).
\end{equation}

Under an orthonormal change of basis, the primitive parameters become
\begin{equation}
\tilde{\boldsymbol{\mu}}_i \in \mathbb{R}^3,
\qquad
\tilde{\boldsymbol{\Sigma}}_i \in \mathbb{R}^{3\times3},
\end{equation}
and the kernel can be written equivalently as
\begin{equation}
T_i(\tilde{\mathbf{x}})=
\left[
1+\frac{1}{\nu_i}
(\tilde{\mathbf{x}}-\tilde{\boldsymbol{\mu}}_i)^\top
\tilde{\boldsymbol{\Sigma}}_i^{-1}
(\tilde{\mathbf{x}}-\tilde{\boldsymbol{\mu}}_i)
\right]^{-\frac{\nu_i+3}{2}}.
\label{eq:appendix_t_ray}
\end{equation}
Since the orthonormal transform has unit Jacobian, the line integral is unchanged. Therefore, the projected contribution of the \(i\)-th primitive is
\begin{equation}
I_i(\hat{\mathbf{x}})
=
\rho_i\int_{-\infty}^{\infty} T_i(\tilde{\mathbf{x}})\,dx_3.
\label{eq:appendix_start_integral}
\end{equation}

\subsection{Quadratic-form decomposition}

Define the precision matrix
\begin{equation}
\tilde{\boldsymbol{\Lambda}}_i
:=
\tilde{\boldsymbol{\Sigma}}_i^{-1}
=
\begin{bmatrix}
\mathbf{A}_i & \mathbf{b}_i \\
\mathbf{b}_i^\top & c_i
\end{bmatrix},
\end{equation}
where \(\mathbf{A}_i\in\mathbb{R}^{2\times2}\), \(\mathbf{b}_i\in\mathbb{R}^{2}\), and \(c_i\in\mathbb{R}\). Since \(\tilde{\boldsymbol{\Sigma}}_i\) is positive definite, \(c_i>0\).

Writing
\begin{equation}
\tilde{\mathbf{x}}-\tilde{\boldsymbol{\mu}}_i
=
\begin{bmatrix}
\hat{\mathbf{x}}-\hat{\boldsymbol{\mu}}_i \\
x_3-\tilde{\mu}_{i,3}
\end{bmatrix},
\end{equation}
the quadratic form can be decomposed by completing the square as
\begin{align}
(\tilde{\mathbf{x}}-\tilde{\boldsymbol{\mu}}_i)^\top
\tilde{\boldsymbol{\Sigma}}_i^{-1}
(\tilde{\mathbf{x}}-\tilde{\boldsymbol{\mu}}_i)
&=
c_i\left(
x_3-\tilde{\mu}_{i,3}
+\frac{\mathbf{b}_i^\top(\hat{\mathbf{x}}-\hat{\boldsymbol{\mu}}_i)}{c_i}
\right)^2 \nonumber\\
&\quad+
(\hat{\mathbf{x}}-\hat{\boldsymbol{\mu}}_i)^\top
\left(
\mathbf{A}_i-\frac{\mathbf{b}_i\mathbf{b}_i^\top}{c_i}
\right)
(\hat{\mathbf{x}}-\hat{\boldsymbol{\mu}}_i).
\label{eq:appendix_decomposed_quad}
\end{align}
We therefore define the effective 2D precision matrix on the detector plane as
\begin{equation}
\hat{\boldsymbol{\Lambda}}_i
=
\mathbf{A}_i-\frac{\mathbf{b}_i\mathbf{b}_i^\top}{c_i},
\qquad
\hat{\boldsymbol{\Sigma}}_i=\hat{\boldsymbol{\Lambda}}_i^{-1}.
\label{eq:appendix_schur}
\end{equation}

\subsection{Closed-form integration along the ray direction}

Define
\begin{equation}
\delta_i(\hat{\mathbf{x}})
=
(\hat{\mathbf{x}}-\hat{\boldsymbol{\mu}}_i)^\top
\hat{\boldsymbol{\Sigma}}_i^{-1}
(\hat{\mathbf{x}}-\hat{\boldsymbol{\mu}}_i),
\label{eq:appendix_delta}
\end{equation}
\begin{equation}
m_i(\hat{\mathbf{x}})
=
\frac{\mathbf{b}_i^\top(\hat{\mathbf{x}}-\hat{\boldsymbol{\mu}}_i)}{c_i}.
\label{eq:appendix_m}
\end{equation}
Substituting Eq.~\eqref{eq:appendix_decomposed_quad} into Eq.~\eqref{eq:appendix_start_integral}, we obtain
\begin{equation}
I_i(\hat{\mathbf{x}})
=
\rho_i
\int_{-\infty}^{\infty}
\left[
1+\frac{\delta_i(\hat{\mathbf{x}})}{\nu_i}
+\frac{c_i}{\nu_i}
\bigl(x_3-\tilde{\mu}_{i,3}+m_i(\hat{\mathbf{x}})\bigr)^2
\right]^{-\frac{\nu_i+3}{2}}
dx_3.
\label{eq:appendix_before_sub}
\end{equation}

Using the change of variable
\begin{equation}
u=
\sqrt{\frac{c_i}{\nu_i+\delta_i(\hat{\mathbf{x}})}}
\bigl(x_3-\tilde{\mu}_{i,3}+m_i(\hat{\mathbf{x}})\bigr),
\label{eq:appendix_u}
\end{equation}
we have
\begin{equation}
dx_3=
\sqrt{\frac{\nu_i+\delta_i(\hat{\mathbf{x}})}{c_i}}\,du.
\end{equation}
Substituting this into Eq.~\eqref{eq:appendix_before_sub} gives
\begin{equation}
I_i(\hat{\mathbf{x}})
=
\rho_i
\left(
1+\frac{\delta_i(\hat{\mathbf{x}})}{\nu_i}
\right)^{-\frac{\nu_i+2}{2}}
\sqrt{\frac{\nu_i}{c_i}}
\int_{-\infty}^{\infty}(1+u^2)^{-\frac{\nu_i+3}{2}}\,du.
\label{eq:appendix_gamma_ready}
\end{equation}

Applying the standard identity
\begin{equation}
\int_{-\infty}^{\infty}(1+u^2)^{-p}\,du
=
\sqrt{\pi}\,
\frac{\Gamma\left(p-\frac12\right)}{\Gamma(p)},
\qquad p>\frac12,
\end{equation}
with \(p=\frac{\nu_i+3}{2}\), we obtain
\begin{equation}
I_i(\hat{\mathbf{x}})
=
\rho_i
\sqrt{\frac{\nu_i\pi}{c_i}}
\frac{
\Gamma\left(\frac{\nu_i+2}{2}\right)
}{
\Gamma\left(\frac{\nu_i+3}{2}\right)
}
\left[
1+\frac{1}{\nu_i}
(\hat{\mathbf{x}}-\hat{\boldsymbol{\mu}}_i)^\top
\hat{\boldsymbol{\Sigma}}_i^{-1}
(\hat{\mathbf{x}}-\hat{\boldsymbol{\mu}}_i)
\right]^{-\frac{\nu_i+2}{2}}.
\label{eq:appendix_projected_t}
\end{equation}

\subsection{Determinant form of the amplitude factor}

From the block determinant identity,
\begin{equation}
|\tilde{\boldsymbol{\Lambda}}_i|
=
c_i\,|\hat{\boldsymbol{\Lambda}}_i|,
\end{equation}
and using \(\tilde{\boldsymbol{\Lambda}}_i=\tilde{\boldsymbol{\Sigma}}_i^{-1}\) and \(\hat{\boldsymbol{\Lambda}}_i=\hat{\boldsymbol{\Sigma}}_i^{-1}\), we obtain
\begin{equation}
\frac{1}{|\tilde{\boldsymbol{\Sigma}}_i|}
=
c_i\frac{1}{|\hat{\boldsymbol{\Sigma}}_i|},
\qquad
c_i=\frac{|\hat{\boldsymbol{\Sigma}}_i|}{|\tilde{\boldsymbol{\Sigma}}_i|}.
\end{equation}
Therefore,
\begin{equation}
\sqrt{\frac{1}{c_i}}
=
\frac{|\tilde{\boldsymbol{\Sigma}}_i|^{1/2}}{|\hat{\boldsymbol{\Sigma}}_i|^{1/2}}.
\end{equation}
Substituting into Eq.~\eqref{eq:appendix_projected_t} yields
\begin{equation}
I_i(\hat{\mathbf{x}})
=
\rho_i\,
\frac{
\Gamma\left(\frac{\nu_i+2}{2}\right)
}{
\Gamma\left(\frac{\nu_i+3}{2}\right)
}
\sqrt{\nu_i\pi}\,
\frac{|\tilde{\boldsymbol{\Sigma}}_i|^{1/2}}{|\hat{\boldsymbol{\Sigma}}_i|^{1/2}}
\left[
1+\frac{1}{\nu_i}
(\hat{\mathbf{x}}-\hat{\boldsymbol{\mu}}_i)^\top
\hat{\boldsymbol{\Sigma}}_i^{-1}
(\hat{\mathbf{x}}-\hat{\boldsymbol{\mu}}_i)
\right]^{-\frac{\nu_i+2}{2}}.
\label{eq:appendix_projected_t_det}
\end{equation}
Hence,
\begin{equation}
C(\nu_i,\tilde{\boldsymbol{\Sigma}}_i)
=
\frac{
\Gamma\left(\frac{\nu_i+2}{2}\right)
}{
\Gamma\left(\frac{\nu_i+3}{2}\right)
}
\sqrt{\nu_i\pi}\,
\frac{|\tilde{\boldsymbol{\Sigma}}_i|^{1/2}}{|\hat{\boldsymbol{\Sigma}}_i|^{1/2}}.
\label{eq:appendix_C_final}
\end{equation}

\subsection{Final projected form}

Combining the above results, the projected contribution of a single primitive is
\begin{equation}
I_i(\hat{\mathbf{x}})
=
\rho_i\,C(\nu_i,\tilde{\boldsymbol{\Sigma}}_i)
\left[
1+\frac{1}{\nu_i}
(\hat{\mathbf{x}}-\hat{\boldsymbol{\mu}}_i)^\top
\hat{\boldsymbol{\Sigma}}_i^{-1}
(\hat{\mathbf{x}}-\hat{\boldsymbol{\mu}}_i)
\right]^{-\frac{\nu_i+2}{2}},
\end{equation}
which matches the projected form in the main text. Summing over all primitives gives the final detector-plane projection.

The key result is that integrating a 3D Student's \(t\)-primitive along the ray direction preserves the Student's \(t\)-form on the detector plane, while the amplitude factor is obtained in closed form as Eq.~\eqref{eq:appendix_C_final}.

\section{Dataset and Preprocessing Details}
\label{app:dataset_details}

This appendix provides additional details on the datasets, preprocessing, projection generation, and sparse-view protocols used in the main paper. It supplements the Datasets and Implementation Details in the experimental setup.

\subsection{Synthetic Dataset}

As described in the main paper, we evaluate TR-GS on a synthetic dataset covering representative CT application scenarios in medical diagnosis, biological research, and industrial inspection. Following the protocol of R$^2$-Gaussian~\cite{Zha2024R2Gaussian}, the synthetic dataset contains 15 cases from three categories: human organs, animals and plants, and artificial objects.

The human-organ category includes \emph{chest}, \emph{foot}, \emph{head}, \emph{jaw}, and \emph{pancreas}, providing anatomically meaningful structures with diverse layouts and boundaries. The animals and plants category contains \emph{beetle}, \emph{bonsai}, \emph{broccoli}, \emph{kingsnake}, and \emph{pepper}, which typically exhibit more irregular geometries and finer details. The artificial-object category consists of \emph{backpack}, \emph{engine}, \emph{present}, \emph{teapot}, and \emph{mount}, and serves as a complementary set with sharper edges, rigid patterns, and repeated structural components.

All raw volumes are normalized to $[0,1]$ and resized to $256 \times 256 \times 256$. We use the TIGRE toolbox~\cite{biguri2016tigre} to simulate X-ray projections with resolution $512 \times 512$ over a full circular trajectory from $0^\circ$ to $360^\circ$.

Unless otherwise specified, the synthetic experiments follow the same sparse-view settings as in the main paper: 12, 18, and 25 views for the main comparisons, and 6 and 50 views for the ablation studies. The default synthetic projection generation uses mixed Poisson--Gaussian noise, where Gaussian noise models electronic readout noise and Poisson noise models photon counting fluctuations. The low-, medium-, and high-noise settings in the robustness study follow the same protocol as in the main paper.

\begin{figure*}[t]
    \centering
    \includegraphics[width=0.9\textwidth]{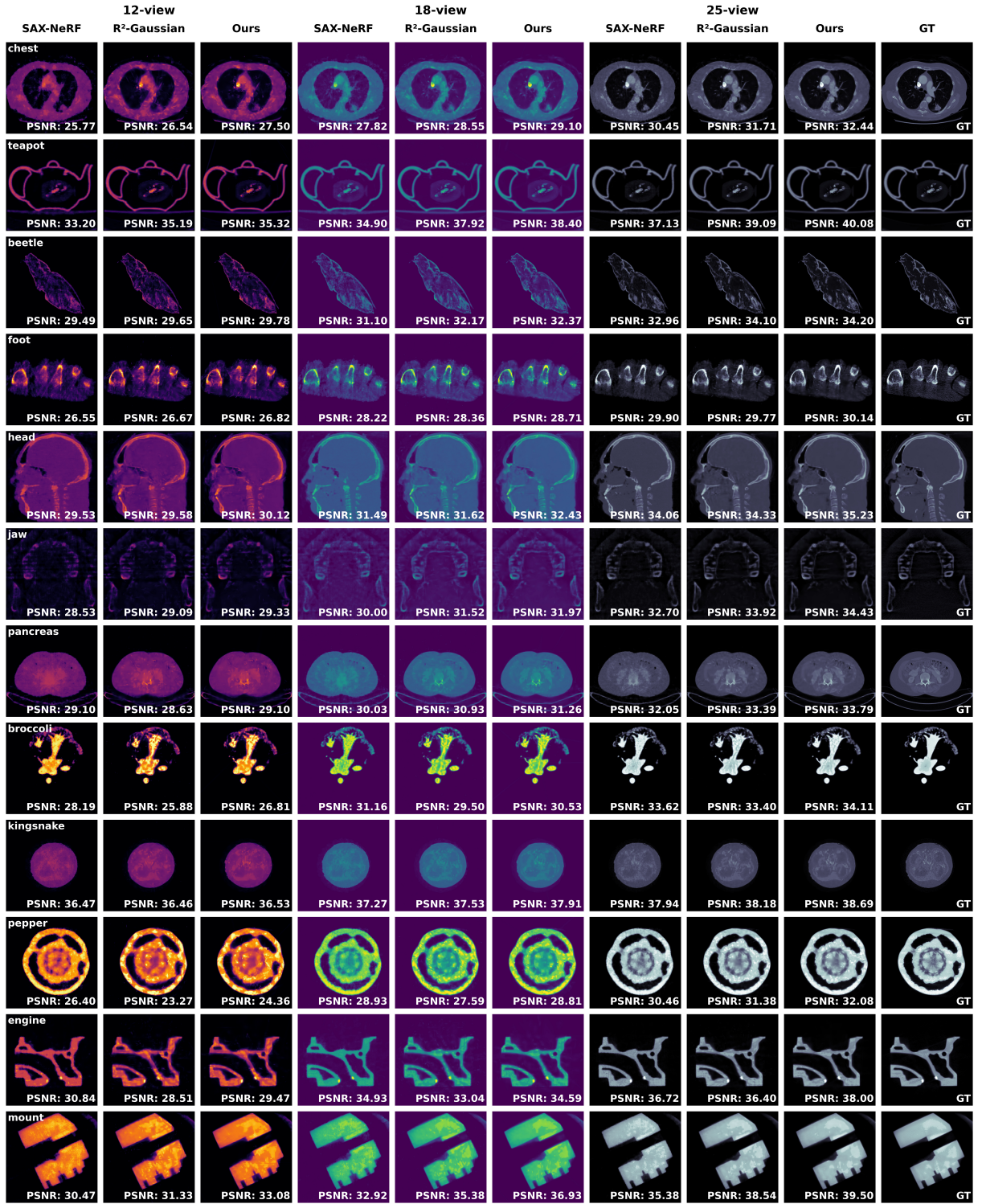}
    \caption{Reconstruction results on the synthetic dataset.}
    \label{fig:synthetic_recon}
\end{figure*}

Figure~\ref{fig:synthetic_recon} shows additional qualitative reconstruction results on the synthetic dataset, complementing the quantitative comparisons in the main paper.

\subsection{Real-World Dataset}

For real-world evaluation, we follow the same real-data protocol as R$^2$-Gaussian~\cite{Zha2024R2Gaussian}, using a public dataset of real captured 2D X-ray projections. The real dataset contains three objects: \emph{pine}, \emph{seashell}, and \emph{walnut}. Although limited in size, these objects provide complementary real acquisition characteristics, including different outer shapes, internal attenuation patterns, and fine structural details.

Each case contains densely sampled projections over a full $0^\circ$--$360^\circ$ rotation. We resize the 2D projections to $560 \times 560$ and normalize intensities to $[0,1]$. Since ground-truth 3D volumes are unavailable, we reconstruct pseudo-ground-truth volumes using FDK with all available views. The target volume resolution is $256 \times 256 \times 256$.

For sparse-view experiments, we subsample projections according to the same settings as in the main paper, namely 12, 18, and 25 views for the main comparisons.

\begin{figure*}[t]
    \centering
    \includegraphics[width=0.9\textwidth]{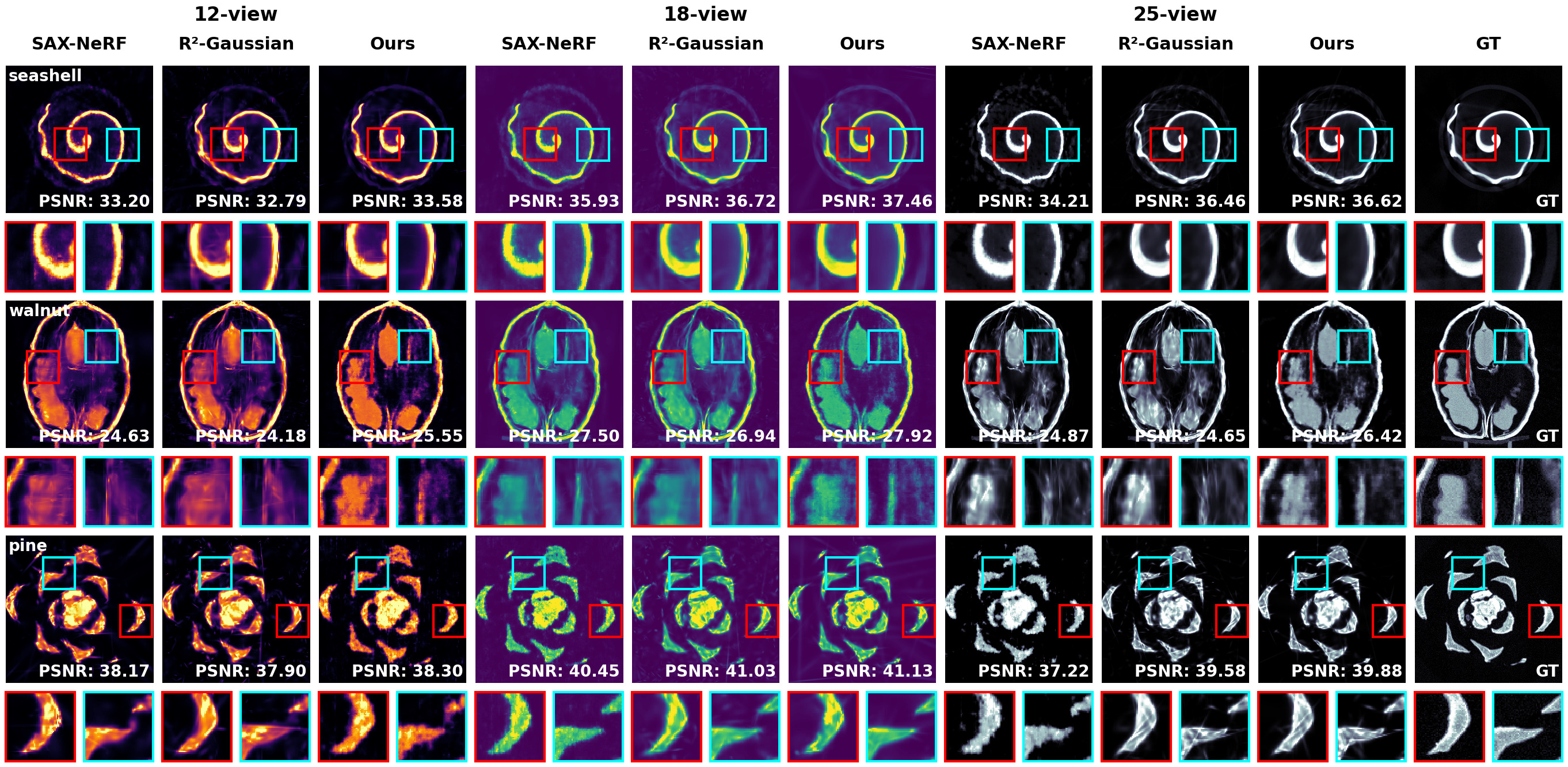}
    \caption{Reconstruction results on the real-world dataset.}
    \label{fig:real_recon}
\end{figure*}

Figure~\ref{fig:real_recon} presents additional qualitative reconstruction results on the real dataset, complementing the quantitative comparisons in the main paper.

\section{Sensitivity Analysis of the Wavelet Module}
\label{app:sensitivity_confidence}

The main paper has already validated the effectiveness of the confidence-guided wavelet regularization through the module-level ablations and reported the sensitivity of the ray-confidence temperature. This appendix therefore focuses on the sensitivity of the remaining control variables in the wavelet module. Unless otherwise specified, all experiments are conducted on the synthetic dataset under the 12-view setting, using the same preprocessing, normalization, training budget, and evaluation protocol as in the main paper. All studies are performed on the full model with all proposed components enabled; only the corresponding internal hyperparameter is varied in each case. We report PSNR\(_{3D}\), SSIM\(_{3D}\), PSNR\(_{2D}\), and SSIM\(_{2D}\).

\subsection{Sensitivity of the Confidence Threshold \(\tau\) in the Wavelet Module}

In the main paper, the confidence threshold \(\tau\) partitions neighboring primitives into high-confidence and low-confidence groups, and therefore controls the routing boundary between enhancement and suppression. We vary \(\tau \in \{0.1, 0.35, 0.5, 0.8\}\), where \(\tau=0.35\) is the default setting.

Table~\ref{tab:wavelet_threshold} shows that the default threshold \(\tau=0.35\) achieves the best overall performance. When \(\tau=0.1\), the enhancement branch is triggered too easily, which weakens the selectivity of the wavelet gating and slightly degrades reconstruction quality. When \(\tau=0.5\), the performance remains close to the default, suggesting that the wavelet module is relatively stable around this range. In contrast, when \(\tau=0.8\), the routing becomes overly conservative and more structurally useful primitives are pushed into suppression, leading to a clearer drop in both 3D and 2D metrics. These results indicate that the wavelet module benefits from a moderate routing boundary rather than an extreme one.

\begin{table}[t]
\centering
\caption{Sensitivity to the confidence threshold \(\tau\) in the confidence-guided wavelet regularization.}
\label{tab:wavelet_threshold}
\setlength{\tabcolsep}{4pt}
\renewcommand{\arraystretch}{1.1}
\scriptsize
\resizebox{\columnwidth}{!}{%
\begin{tabular}{lcccc}
\toprule
\multicolumn{5}{c}{Synthetic, 12-view} \\
\midrule
\(\tau\) & PSNR\(_{3D}\) \(\uparrow\) & SSIM\(_{3D}\) \(\uparrow\) & PSNR\(_{2D}\) \(\uparrow\) & SSIM\(_{2D}\) \(\uparrow\) \\
\midrule
0.10 & 30.95 & 0.870 & 40.74 & 0.968 \\
0.35 & 31.05 \(\checkmark\) & 0.874 \(\checkmark\) & 40.79 \(\checkmark\) & 0.974 \(\checkmark\) \\
0.50 & 30.99 & 0.870 & 40.78 & 0.972 \\
0.80 & 30.94 & 0.866 & 40.65 & 0.969 \\
\bottomrule
\end{tabular}%
}
\end{table}

\subsection{Sensitivity of the Enhancement Strength in the Wavelet Module}

After the routing is determined by \(\tau\), the enhancement branch further scales high-confidence high-frequency responses according to an enhancement factor. We fix \(\tau=0.35\) and vary the enhancement factor in \(\{1.5, 2.0, 4.0, 8.0\}\).

As shown in Table~\ref{tab:wavelet_alpha}, the default enhancement factor \(1.5\) achieves the best overall performance among the tested settings, confirming that moderate confidence-guided high-frequency enhancement is beneficial for recovering structural details blurred by the heavy-tailed \(t\)-primitive representation. Further increasing it to 2.0 does not lead to a uniform gain across all metrics, indicating that the performance has already entered a plateau regime. When the enhancement factor is increased to 4.0 and 8.0, the quantitative performance begins to degrade, especially in 2D consistency and SSIM\(_{2D}\), suggesting that excessive amplification tends to over-enhance local high-frequency responses rather than improve meaningful detail recovery. Overall, the results indicate a stable operating range around the default setting.

\begin{table}[t]
\centering
\caption{Sensitivity to the enhancement strength in the confidence-guided wavelet module.}
\label{tab:wavelet_alpha}
\setlength{\tabcolsep}{4pt}
\renewcommand{\arraystretch}{1.1}
\scriptsize
\resizebox{\columnwidth}{!}{%
\begin{tabular}{lcccc}
\toprule
\multicolumn{5}{c}{Synthetic, 12-view} \\
\midrule
Enhancement factor & PSNR\(_{3D}\) \(\uparrow\) & SSIM\(_{3D}\) \(\uparrow\) & PSNR\(_{2D}\) \(\uparrow\) & SSIM\(_{2D}\) \(\uparrow\) \\
\midrule
1.50 & 31.05 \(\checkmark\) & 0.874 \(\checkmark\) & 40.79 \(\checkmark\) & 0.974 \(\checkmark\) \\
2.00 & 31.03 & 0.873 & 40.78 & 0.972 \\
4.00 & 31.02 & 0.870 & 40.71 & 0.972 \\
8.00 & 30.96 & 0.870 & 40.69 & 0.968 \\
\bottomrule
\end{tabular}%
}
\end{table}

\subsection{Summary}

Overall, these experiments show that the default settings used in the main paper lie in a stable operating regime under the synthetic 12-view setting. In the wavelet module, \(\tau\) controls the enhancement--suppression routing and should remain in a moderate range, while the enhancement factor should be strong enough to recover fine structures but not so large as to over-amplify local oscillations. Therefore, the results in this appendix should be viewed as a sensitivity analysis of the internal control variables of the proposed mechanisms, rather than an additional module-level ablation.

\end{document}